\documentclass{article}
\usepackage{times}
\usepackage{graphicx}
\usepackage{subfigure} 
\usepackage{natbib}
\usepackage{epsfig}
\usepackage{epstopdf} 
\usepackage{enumitem}
\usepackage{amsmath}
\usepackage{amssymb}
\usepackage{amsfonts}
\usepackage{amsthm}
\usepackage{subfigure}
\usepackage{verbatim}
\newcommand{\argmin}{\operatornamewithlimits{argmin}}
\newcommand{\circR}{\operatornamewithlimits{circ}}
\newcommand{\sign}{\operatornamewithlimits{sign}}
\newcommand{\Tr}{\operatornamewithlimits{Tr}}

\newtheorem{proposition}{Proposition}

\usepackage[accepted]{icml2014}
\icmltitlerunning{Circulant Binary Embedding}

\begin{document} 
\twocolumn[
\icmltitle{Circulant Binary Embedding}
\icmlauthor{Felix X. Yu$^1$}{yuxinnan@ee.columbia.edu}
\icmlauthor{Sanjiv Kumar$^2$}{sanjivk@google.com}
\icmlauthor{Yunchao Gong$^3$}{yunchao@cs.unc.edu}
\icmlauthor{Shih-Fu Chang$^1$}{sfchang@ee.columbia.edu}
\icmladdress{$^1$Columbia University, New York, NY 10027}
\vspace{-0.3cm}
\icmladdress{$^2$Google Research, New York, NY 10011}
\vspace{-0.3cm}
\icmladdress{$^3$University of North Carolina at Chapel Hill, Chapel Hill, NC 27599}
\vskip 0.3in
]

\begin{abstract}
Binary embedding of high-dimensional data requires long codes to preserve the discriminative power of the input space. Traditional binary coding methods often suffer from very high computation and storage costs in such a scenario. To address this problem, we propose Circulant Binary Embedding (CBE) which generates binary codes by projecting the data with a circulant matrix. The circulant structure enables the use of Fast Fourier Transformation to speed up the computation. 
Compared to methods that use unstructured matrices, the proposed method improves the time complexity from $\mathcal{O}(d^{2})$ to $\mathcal{O}(d\log{d})$, and the space complexity from $\mathcal{O}(d^2)$ to $\mathcal{O}(d)$ where $d$ is the input dimensionality.
We also propose a novel time-frequency alternating optimization to learn data-dependent circulant projections, which alternatively minimizes the objective in original and Fourier domains. 
We show by extensive experiments that the proposed approach gives much better performance than the state-of-the-art approaches for fixed time, and provides much faster computation with no performance degradation for fixed number of bits.
\end{abstract}

\section{Introduction}
\label{sec:intro}
Embedding input data in binary spaces  is becoming popular for efficient retrieval and learning on massive data sets \cite{li2011hashing, gonglearning, raginsky2009locality, gong2012angular, liu2011hashing}.
Moreover, in a large number of application domains such as computer vision, biology and finance, data is typically high-dimensional. When representing such high dimensional data by binary codes, it has been shown that long codes are required in order to achieve good performance. In fact, the required number of bits is $\mathcal{O}(d)$, where $d$ is the input dimensionality \cite{li2011hashing, gonglearning, sanchez2011high}. The goal of binary embedding is to well approximate the input distance as Hamming distance so that efficient learning and retrieval can happen directly in the binary space. It is important to note that another related area called \textit{hashing} is a special case with slightly different goal: creating hash tables such that points that are similar fall in the same (or nearby) bucket with high probability. In fact, even in hashing, if high accuracy is desired, one typically needs to use hundreds of hash tables involving tens of thousands of bits.

\vspace{-0.1cm}

Most of the existing linear binary coding approaches generate the binary code by applying a  projection matrix, followed by a binarization step. Formally, given a data point, $\mathbf{x} \in \mathbb{R}^{d}$, the $k$-bit binary code, $h(\mathbf{x}) \in \{+1, -1\}^k$ is generated simply as  
\vspace{-0.2cm}
\begin{equation}
h(\mathbf{x}) = \text{sign} (\mathbf{R} \mathbf{x}),
\label{eq:lsh_intro}
\end{equation}
where $\mathbf{R} \in \mathbb{R}^{k \times d}$, and $\text{sign}(\cdot)$ is a binary map which returns element-wise sign\footnote{A few methods transform the linear projection via a nonlinear map before taking the sign \cite{weiss2008spectral, raginsky2009locality}.}. Several techniques have been proposed to generate the projection matrix randomly without taking into account the input data \cite{charikar2002similarity, raginsky2009locality}. These methods are very popular due to their simplicity but often fail to give the best performance due to their inability to adapt the codes with respect to the input data.  Thus, a number of data-dependent techniques have been proposed with different optimization criteria such as reconstruction error \cite{Kulislearningto}, data dissimilarity \cite{Norouzi11, weiss2008spectral}, ranking loss \cite{norouzi2012Hamming}, quantization error after PCA \cite{gongiterative}, 
and pairwise misclassification \cite{wang2010sequential}. These methods are shown to be effective for learning compact codes for relatively low-dimensional data. 
However, the $\mathcal{O}(d^2)$  computational and space costs prohibit them from being applied to learning long codes for high-dimensional data. For instance, to generate $\mathcal{O}(d)$-bit binary codes for data with $d \sim $1M, a huge projection matrix will be required needing TBs of memory, which is not practical\footnote{In principle, one can generate the random entries of the matrix on-the-fly (with fixed seeds) without needing to store the matrix. But this will increase the computational time even further.}.  

\vspace{-0.15cm}

In order to overcome these computational challenges, \citet{gonglearning} proposed a bilinear projection based coding method for high-dimensional data. It reshapes the input vector $\mathbf{x}$ into a matrix $\mathbf{Z}$, and applies a bilinear projection to get the binary code:
\begin{eqnarray}
\vspace{-0.45cm}
h(\mathbf{x}) = \text{sign} (\mathbf{R}_1^T \mathbf{Z} \mathbf{R}_2).
\vspace{-0.45cm}
\end{eqnarray}
When the shapes of $\mathbf{Z}, \mathbf{R}_1, \mathbf{R}_2$ are chosen appropriately, the method has time and space complexity of  $\mathcal{O}(d^{1.5})$ and  $\mathcal{O}(d)$, respectively. 
Bilinear codes make it feasible to work with datasets with very high dimensionality and have shown good results in a variety of tasks. 

\vspace{-0.1cm}

In this work, we propose a novel Circulant Binary Embedding (CBE) technique which is even faster than the bilinear coding. It is achieved by imposing a circulant structure on the projection matrix $\mathbf{R}$ in (\ref{eq:lsh_intro}). This special structure allows us to use Fast Fourier Transformation (FFT) based techniques, which have been extensively used in signal processing. The proposed method further reduces the time complexity to $\mathcal{O}(d \log d)$, enabling efficient binary embedding for very high-dimensional data\footnote{One could in principal use other structured matrices like Hadamard matrix along with a sparse random Gaussian matrix to achieve fast projection as was done in fast Johnson-Lindenstrauss transform\cite{ailon2006approximate, dasgupta2011fast}, but it is still slower than circulant projection and needs more space.}.
Table \ref{table:methods} compares the time and space complexity for different methods. This work makes the following contributions:
\vspace{-0.2cm}
\begin{itemize}[noitemsep]
\item We propose the circulant binary embedding method, which has space complexity $\mathcal{O}(d)$ and time complexity $\mathcal{O}(d \log d)$ (Section \ref{sec:cbe}, \ref{sec:rand}).
\item We propose to learn the data-dependent circulant projection matrix by a novel and efficient time-frequency alternating optimization, which alternatively optimizes the objective in the original and frequency domains (Section \ref{sec:opt}).
\item Extensive experiments show that, compared to the state-of-the-art, the proposed method improves the result dramatically for a fixed time cost, and provides much faster computation with no performance degradation for a fixed number of bits (Section \ref{sec:exp}).
\end{itemize}

\begin{small}
\begin{table}
\begin{center}
\small
\begin{tabular}{|l|l|l|l|}
\hline Method & Time   & Space  & Time (Learning)\\ 
\hline   Full projection    & $\mathcal{O}(d^2)$ & $\mathcal{O}(d^2)$ & $\mathcal{O}(nd^3)$\\ 
\hline   Bilinear proj. & $\mathcal{O}(d^{1.5})$ &  $\mathcal{O}(d)$ &    $\mathcal{O}(nd^{1.5})$\\ 
\hline   Circulant proj. & $\mathcal{O}(d\log{d})$ & $\mathcal{O}(d)$  & $\mathcal{O}(nd\log{d})$\\ 
\hline 
\end{tabular}
\end{center}
\vspace{-0.3cm}
\caption{Comparison of the proposed method (Circulant proj.) with other methods for generating long codes (code dimension $k$ comparable to input dimension $d$). $n$ is the number of instances used for learning data-dependent projection matrices. }
\label{table:methods}
\vspace{-0.3cm}
\end{table}
\end{small}

\section{Circulant Binary Embedding (CBE)}
\label{sec:cbe}

A circulant matrix $\mathbf{R} \in \mathbb{R}^{d \times d}$ is a matrix defined by a vector $\mathbf{r}  = (r_0, r_2, \cdots,  r_{d-1})^T$ \cite{gray2006toeplitz}\footnote{The circulant matrix is sometimes equivalently defined by ``circulating'' the rows instead of the columns.}. 
\begin{align}
\mathbf{R} = \circR(\mathbf{r}) :=
\begin{bmatrix}
r_0     & r_{d-1} & \dots  & r_{2} & r_{1}  \\
r_{1} & r_0    & r_{d-1} &         & r_{2}  \\
\vdots  & r_{1}& r_0    & \ddots  & \vdots   \\
r_{d-2}  &        & \ddots & \ddots  & r_{d-1}   \\
r_{d-1}  & r_{d-2} & \dots  & r_{1} & r_{0}
\end{bmatrix}.
\label{eq:cir}
\end{align}

Let $\mathbf{D}$ be a diagonal matrix with each diagonal entry being a Bernoulli  variable ($\pm 1$ with probability 1/2).
For $\mathbf{x} \in \mathbb{R}^d$, its $d$-bit Circulant Binary Embedding (CBE) with $\mathbf{r} \in \mathbb{R}^{d}$ is defined as:
\begin{align}
\vspace{-0.4cm}
h(\mathbf{x}) = \text{sign} (\mathbf{R} \mathbf{D}  \mathbf{x}),
\label{eq:def}
\vspace{-0.2cm}
\end{align}
where $\mathbf{R} = \circR(\mathbf{r})$.
The $k$-bit ($k<d$) CBE is defined as the first $k$ elements of $h(\mathbf{x})$.
The need for such a $\mathbf{D}$ is discussed in Section \ref{sec:rand}. Note that applying $\mathbf{D}$ to $\mathbf{x}$ is equivalent to applying random sign flipping to each dimension of $\mathbf{x}$. Since sign flipping can be carried out as a preprocessing step for each input $\mathbf{x}$, here onwards for simplicity we will drop explicit mention of $\mathbf{D}$. Hence the binary code is given as $h(\mathbf{x}) = \text{sign} (\mathbf{R} \mathbf{x})$.
\vspace{-0.1cm}

The main advantage of circulant binary embedding is its ability to use Fast Fourier Transformation (FFT) to speed up the computation. 

\begin{proposition}
For $d$-dimensional data, CBE has space complexity $\mathcal{O}(d)$, and time complexity $\mathcal{O}(d \log d)$.
\label{prop:time}
\end{proposition}
\vspace{-0.1cm}

Since a circulant matrix is defined by a single column/row, clearly the storage needed is $\mathcal{O}(d)$. 
Given a data point $\mathbf{x}$, the $d$-bit CBE can be efficiently computed as follows. 
Denote $\circledast$ as operator of circulant convolution. Based on the definition of circulant matrix,
\begin{align}
\mathbf{R} \mathbf{x} = \mathbf{r} \circledast \mathbf{x}.
\vspace{-0.2cm}
\end{align}
The above can be computed based on Discrete Fourier Transformation (DFT), for which fast algorithm (FFT) is available. 
%
The DFT of a vector $\mathbf{t} \in \mathbb{C}^{d}$ is a $d$-dimensional vector with each element defined as 
\vspace{-0.1cm}
\begin{equation}
\mathcal{F}(\mathbf{t})_l = \sum_{m=0}^{d-1} t_n \cdot e^{-i 2 \pi l m /d}, l = 0, \cdots, d-1.
\end{equation}
\vspace{-0.15cm}

The above can be expressed equivalently in a matrix form as
\begin{equation}
\mathcal{F}(\mathbf{t}) = \mathbf{F}_d \mathbf{t},
\end{equation}
where $\mathbf{F}_d$ is the $d$-dimensional DFT matrix.
Let $\mathbf{F}_d^H$ be the conjugate transpose of $\mathbf{F}_d$.
It is easy to show that $\mathbf{F}_d^{-1} = (1/d) \mathbf{F}_d^H$.
Similarly, for any $\mathbf{t} \in \mathbb{C}^d$, the Inverse Discrete Fourier Transformation (IDFT) is defined as 
\begin{equation}
\mathcal{F}^{-1}(\mathbf{t}) = (1/d) \mathbf{F}_d^H\mathbf{t}.
\end{equation}
\vspace{-0.5cm}

Since the convolution of two signals in their original domain is equivalent to the hadamard product in their frequency domain \cite{oppenheim1999discrete},
\begin{align}
\mathcal{F} (\mathbf{R} \mathbf{x}) =  \mathcal{F}({\mathbf{r} }) \circ \mathcal{F}(\mathbf{x}).
\end{align}
Therefore, 
\vspace{-0.1cm}
\begin{align}
h (\mathbf{x}) = \text{sign} \left( \mathcal{F}^{-1} ( \mathcal{F}({\mathbf{r} }) \circ \mathcal{F}(\mathbf{x})) \right).
\vspace{-0.2cm}\end{align}
For $k$-bit CBE, $k<d$, we only need to pick the first $k$ bits of $h (\mathbf{x})$. As DFT and IDFT can be efficiently computed in $\mathcal{O}(d \log{d})$ with FFT \cite{oppenheim1999discrete}, generating CBE has time complexity $\mathcal{O}(d \log{d})$.

\vspace{-0.1cm}

\section{Randomized Circulant Binary Embedding}
\label{sec:rand}
\vspace{-0.1cm}

A simple way to obtain CBE is by generating the elements of $\mathbf{r}$ in (\ref{eq:cir}) independently from the standard normal distribution $\mathcal{N}(0,1)$. We call this method randomized CBE (CBE-rand).
A desirable property of any embedding method is its ability to approximate input distances in the embedded space. Suppose $\mathcal{H}_k(\mathbf{x}_1, \mathbf{x}_2) $ is the normalized Hamming distance between $k$-bit codes of a pair of points $\mathbf{x}_1, \mathbf{x}_2 \in \mathbb{R}^d$:
\begin{equation}
\mathcal{H}_k(\mathbf{x}_1, \mathbf{x}_2)\! =\! 
\frac{1}{k} \!\sum_{i=0}^{k-1} \!
\big|\!
\sign(\mathbf{R}_{i\cdot}\mathbf{x}_1) \!-\! \sign(\mathbf{R}_{i\cdot}\mathbf{x}_2)
\big|/2,
\end{equation}
and $\mathbf{R}_{i\cdot}$ is the $i$-th row of $\mathbf{R}$, $\mathbf{R} = \circR(\mathbf{r})$.
If $\mathbf{r}$ is sampled from $\mathcal{N}(0,1)$, from \cite{charikar2002similarity},   
\begin{equation}
\textbf{Pr} \left(\text{sign} (\mathbf{r}^T \mathbf{x}_1) \ne \text{sign} (\mathbf{r}^T \mathbf{x}_2)  \right)
 =  \theta/\pi,
 \label{eq:lsh}
\end{equation}
where $ \theta$ is the angle between $\mathbf{x}_1$ and $\mathbf{x}_2$. Since all the vectors that are circulant variants of $\mathbf{r}$ also follow the same distribution, it is easy to see that
\begin{align}
& \textbf{E}(\mathcal{H}_k(\mathbf{x}_1, \mathbf{x}_2) ) =  \theta/\pi. \label{eq:e}
\end{align}

\vspace{-0.1cm}

\begin{figure}
\centering
\vspace{-0.2cm}
\subfigure[$\theta = \pi/12$]
{\includegraphics[width=4.2cm]{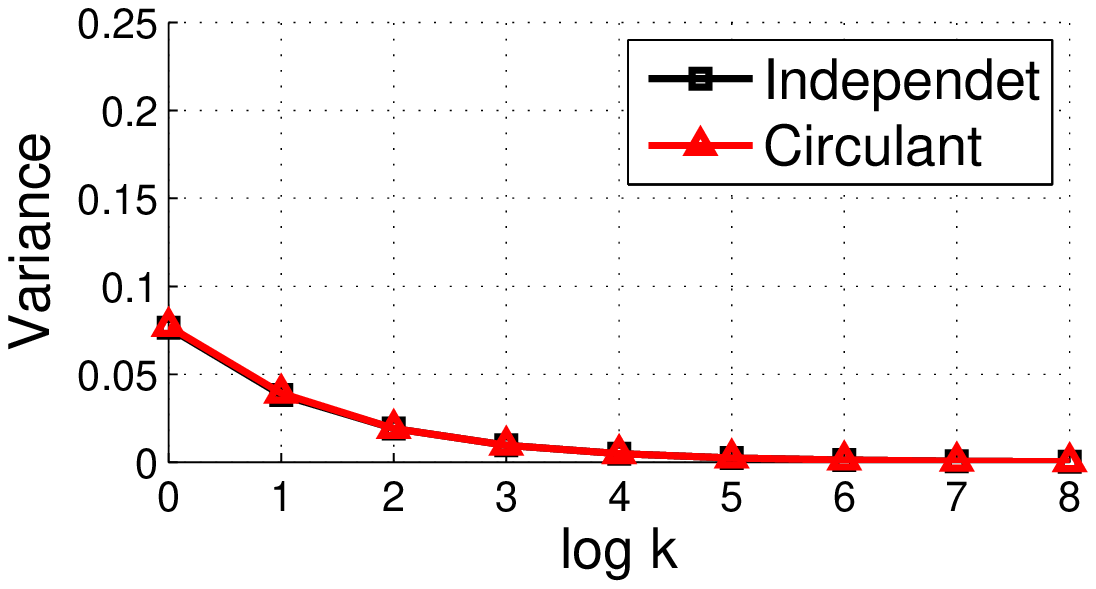}}\hspace{-0.3cm}
\subfigure[$\theta = \pi/6$]
{\includegraphics[width=4.2cm]{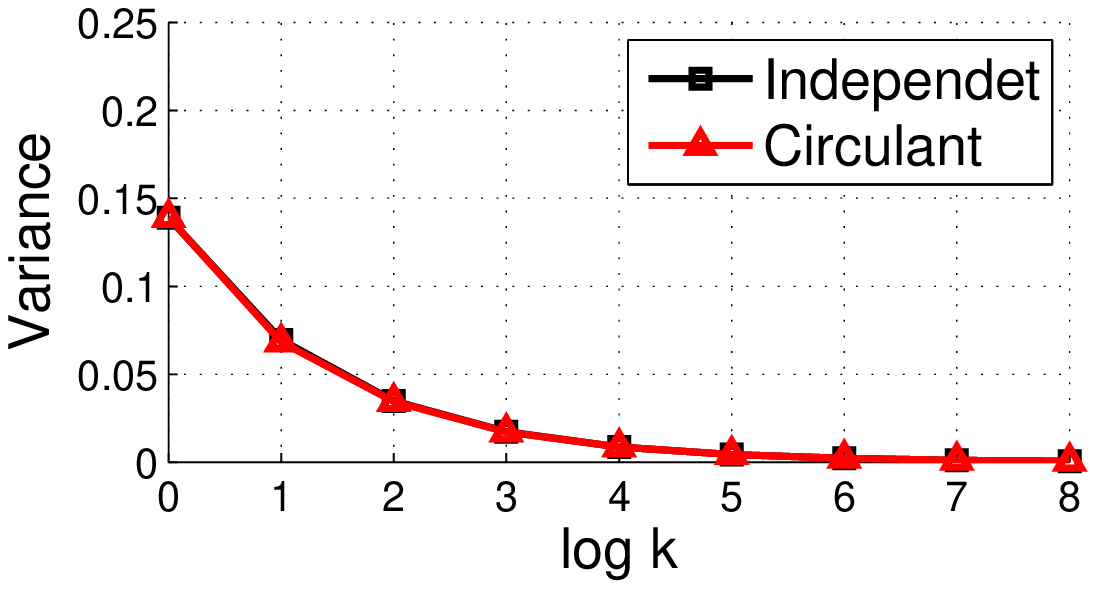}} \\ \vspace{-0.2cm}
\subfigure[$\theta = \pi/3$]
{\includegraphics[width=4.2cm]{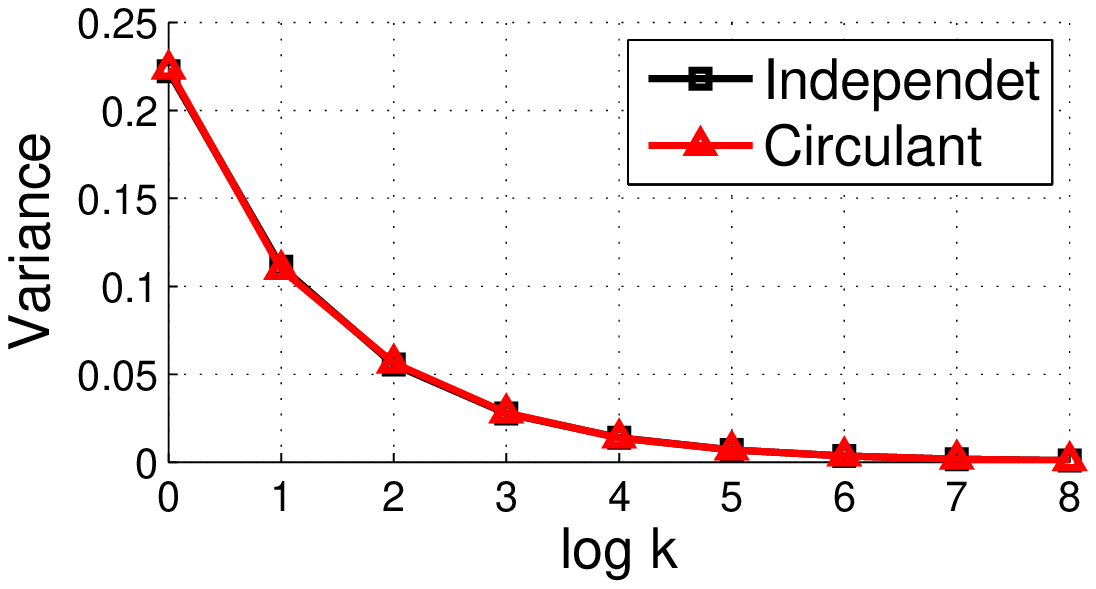}}\hspace{-0.3cm}
\subfigure[$\theta = \pi/2$]
{\includegraphics[width=4.2cm]{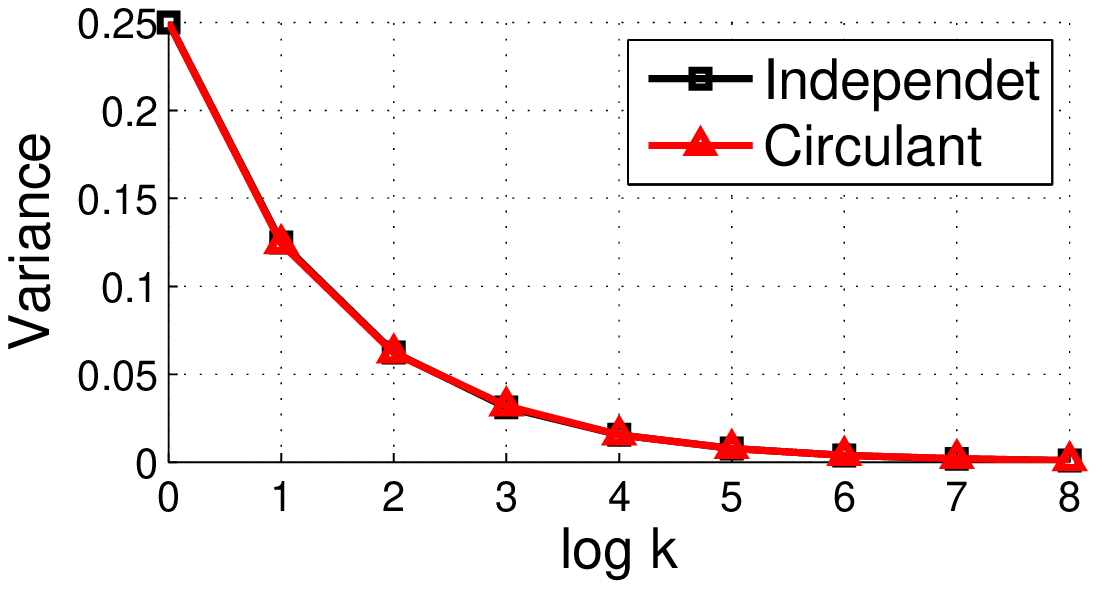}}
\vspace{-0.4cm}
\caption{The \emph{analytical} variance of normalized hamming distance of independent bits as in (\ref{eq:var}), and the \emph{sample} variance of normalized hamming distance of circulant bits, as a function of angle between points ($\theta$) and number of bits ($k$). The two curves overlap.}
\label{fig:variance}
\vspace{-1cm}
\end{figure}

For the sake of discussion, 
if $k$ projections, \emph{i.e.}, first $k$ rows of $\mathbf{R}$, were generated independently, 
it is easy to show that the variance of $\mathcal{H}_k(\mathbf{x}_1, \mathbf{x}_2) $ will be
 \begin{align}
& \textbf{Var}(\mathcal{H}_k(\mathbf{x}_1, \mathbf{x}_2)) = \theta(\pi - \theta)/k\pi^2.\label{eq:var}
\end{align}
Thus, with more bits (larger $k$), the normalized hamming distance will be close to the expected value, with lower variance. In other words, the normalized hamming distance approximately preserves the angle\footnote{In this paper, we consider the case that the data points are $\ell_2$ normalized. Therefore the cosine distance, \emph{i.e.}, 1 - $\cos(\theta)$, is equivalent to the $l_2$ distance.}.
Unfortunately in CBE, the projections are the rows of $\mathbf{R} = \circR(\mathbf{r})$, which are not independent. This makes it hard to derive the variance analytically. 
To better understand CBE-rand, we run simulations to compare the analytical variance of normalized hamming distance of independent projections (\ref{eq:var}), and the \emph{sample} variance of normalized hamming distance of circulant projections in Figure \ref{fig:variance}. 
For each $\theta$ and $k$, we randomly generate $\mathbf{x}_1, \mathbf{x}_2 \in \mathbb{R}^d$ such that their angle is $\theta$\footnote{This can be achieved by extending the 2D points $(1,0)$, $(\cos\theta, \sin\theta)$ to $d$-dimension, and performing a random orthonormal rotation, which can be formed by the Gram-Schmidt process on random vectors.}.
We then generate $k$-dimensional code with CBE-rand, and compute the hamming distance. The variance is estimated by applying CBE-rand 1,000 times. We repeat the whole process 1,000 times, and compute the averaged variance. 
Surprisingly, the curves of ``Independent'' and ``Circulant'' variances are almost indistinguishable. This means that bits generated by CBE-rand are generally as good as the independent bits for angle preservation. An intuitive explanation is that for the circulant matrix, though all the rows are dependent, circulant shifting one or multiple elements will in fact result in very different projections in most cases. 
We will later show in experiments on real-world data that CBE-rand and Locality Sensitive Hashing (LSH)\footnote{Here, by LSH we imply the binary embedding using $\mathbf{R}$ such that all the rows of $\mathbf{R}$ are sampled iid. With slight abuse of notation, we still call it ``hashing'' following \cite{charikar2002similarity}.} has almost identical performance (yet CBE-rand is significantly faster) (Section \ref{sec:exp}).

\vspace{-0.1cm}

Note that the distortion in input distances after circulant binary embedding comes from two sources: circulant projection, and binarization. 
For the circulant projection step, recent works have shown that the Johnson-Lindenstrauss-type lemma holds with a slightly worse bound on the number of projections needed to preserve the input distances with high probability \cite{hinrichs2011johnson, zhang2013new, vybiral2011variant, krahmer2011new}. These works also show that before applying the circulant projection, an additional step of randomly flipping the signs of input dimensions is necessary\footnote{For each dimension, whether the sign needs to be flipped is predetermined by a $(p=0.5)$ Bernoulli variable.}. To show why such a step is required, let us consider the special case when $\mathbf{x}$ is an all-one vector, $\mathbf{1}$. The circulant projection with $\mathbf{R} = \circR(\mathbf{r})$ will result in a vector with all elements to be $\mathbf{r}^T \mathbf{1}$. When $\mathbf{r}$ is independently drawn from $\mathcal{N}(0,1)$, this will be close to 0, and the norm cannot be preserved.
Unfortunately the Johnson-Lindenstrauss-type results do not generalize to the distortion caused by the binarization step. 

\vspace{-0.1cm}

One problem with the randomized CBE method is that it does not utilize the underlying data distribution while generating the matrix $\mathbf{R} $. In the next section, we propose to learn $\mathbf{R}$ in a data-dependent fashion, to minimize the distortions due to circulant projection and binarization.
 
\vspace{-0.2cm}

\section{Learning Circulant Binary Embedding}
\label{sec:opt}

We propose data-dependent CBE (CBE-opt), by optimizing the projection matrix with a novel time-frequency alternating optimization. 
We consider the following objective function in learning the $d$-bit CBE. The extension of learning $k < d$ bits will be shown in Section \ref{subsec:k}.
\begin{align}
\vspace{-0.6cm}
\argmin_{\mathbf{B}, \mathbf{r}}   \quad\!\!  &|| \mathbf{B} - \mathbf{X} \mathbf{R}^T ||_F^2 +  \lambda|| \mathbf{R} \mathbf{R}^T - \mathbf{I}||_F^2 \label{eq:obj}\\
\text{s.t.}  \quad &  \mathbf{R} = \circR(\mathbf{r}), \nonumber
\vspace{-0.6cm}
\end{align}
where $\mathbf{X} \in  \mathbb{R}^{n \times d}$, is the data matrix containing $n$ training points: $\mathbf{X} = [\mathbf{x}_0, \cdots, \mathbf{x}_{n-1}]^T$, and
  $\mathbf{B} \in  \{-1,1\}^{n \times d}$ is the corresponding binary code matrix.\footnote{If the data is $\ell_2$ normalized, we can set $\mathbf{B} \in  \{-1/\sqrt{d},1/\sqrt{d} \}^{n \times d}$ to make $\mathbf{B}$ and $\mathbf{X}\mathbf{R}^T$ more comparable. This does not empirically influence the performance.}

In the above optimization, the first term minimizes distortion due to binarization. The second term tries to make the projections (rows of $\mathbf{R}$, and hence the corresponding bits) as uncorrelated as possible. In other words, this helps to reduce the redundancy in the learned code. 
%
If $\mathbf{R}$ were to be an orthogonal matrix, the second term will vanish and the optimization would find the best rotation such that the distortion due to binarization is minimized. 
However, when $\mathbf{R}$ is a circulant matrix, $\mathbf{R}$, in general, will not be orthogonal. 
Similar objective has been used in previous works including \cite{gongiterative, gonglearning} and \cite{wang2010sequential}.

\subsection{The Time-Frequency Alternating Optimization}
The above is a combinatorial optimization problem, for which an optimal solution is hard to find. In this section we propose a novel approach to efficiently find a local solution.
The idea is to alternatively optimize the objective by fixing $\mathbf{r}$, and $\mathbf{B}$, respectively. 
For a fixed $\mathbf{r}$, optimizing $\mathbf{B}$ can be easily performed in the input domain (``time'' as opposed to ``frequency''). 
For a fixed $\mathbf{B}$, the circulant structure of $\mathbf{R}$ makes it difficult to optimize the objective in the input domain. Hence we propose a novel method, by optimizing $\mathbf{r}$ in the frequency domain based on DFT. This leads to a very efficient procedure. 
%

\textbf{For a fixed $\mathbf{r}$}.\! 
The objective is independent on each element of $\mathbf{B}$. Denote $B_{ij}$ as the element of the $i$-th row and $j$-th column of $\mathbf{B}$. It is easy to show that $\mathbf{B}$ can be updated as:
\begin{align}
\vspace{-1cm}
& B_{ij} = \begin{cases}
   1 & \text{if } \mathbf{R}_{j\cdot}\mathbf{x}_i \geq 0 \\
   -1 & \text{if } \mathbf{R}_{j\cdot}\mathbf{x}_i < 0
  \end{cases}, \\
&  i = 0, \cdots, n-1. \quad j = 0, \cdots, d-1. \nonumber
\vspace{-0.3cm}
\end{align}
\textbf{For a fixed $\mathbf{B}$}. Define $\tilde{\mathbf{r}}$ as the DFT of the circulant vector $\tilde{\mathbf{r}} := \mathcal{F}(\mathbf{r})$. Instead of solving $\mathbf{r}$ directly, we propose to solve $\tilde{\mathbf{r}}$, from which $\mathbf{r}$ can be recovered by IDFT. 

Key to our derivation is the fact that DFT projects the signal to a set of orthogonal basis. Therefore the $\ell_2$ norm can be preserved. Formally, according to Parseval's theorem , for any $\mathbf{t} \in \mathbb{C}^d$ \cite{oppenheim1999discrete}, 
\begin{equation}
|| \mathbf{t}||_2^2 = (1/d) ||\mathcal{F}(\mathbf{t})||_2^2. \nonumber
\vspace{-0.2cm}
\end{equation}

Denote $\text{diag}(\cdot)$ as the diagonal matrix formed by a vector. Denote $\Re(\cdot)$ and $\Im(\cdot)$ as the real and imaginary parts, respectively. We use $\mathbf{B}_{i\cdot}$ to denote the $i$-th row of $\mathbf{B}$. With complex arithmetic, the first term in (\ref{eq:obj}) can be expressed in the frequency domain as:

\vspace{-0.5cm}
\begin{small}
\begin{align}
& || \mathbf{B} - \mathbf{X} \mathbf{R}^T ||_F^2 = \frac{1}{d}\sum_{i = 0}^{n-1} ||\mathcal{F}(\mathbf{B}^T_{i \cdot} -  \mathbf{R} \mathbf{x}_i) ||_2^2 \label{eq:bxr}
\\ 
= & \frac{1}{d} \! \sum_{i = 0}^{n-1} \!\! ||\mathcal{F} (\mathbf{B}^T_{i \cdot}) \!\!- \!\! \tilde{\mathbf{r}} \!\circ\! \mathcal{F}(\mathbf{x}_i) ||_2^2 \!=\! \frac{1}{d}\!\sum_{i = 0}^{n-1}\! \!||\mathcal{F} (\mathbf{B}^T_{i \cdot}) \!\!-\!\! \text{diag}(\!{\mathcal{F}(\mathbf{x}_i)} ) \tilde{\mathbf{r}} ||_2^2 \nonumber \\
= & \frac{1}{d}\!\sum_{i = 0}^{n-1}\! \left(\mathcal{F} (\mathbf{B}^T_{i \cdot}) \!-\! \text{diag}({\mathcal{F}(\mathbf{x}_i)} ) \tilde{\mathbf{r}} \right)\!^T\! 
\left(\mathcal{F} (\mathbf{B}^T_{i \cdot}) \!-\! \text{diag}({\mathcal{F}(\mathbf{x}_i)} ) \tilde{\mathbf{r}} \right) \nonumber \\
= & \frac{1}{d} \Big[ \Re(\tilde{\mathbf{r}})\!^T \mathbf{M} \Re(\tilde{\mathbf{r}}) \! + \! \Im(\tilde{\mathbf{r}})\!^T \mathbf{M} \Im(\tilde{\mathbf{r}}) 
\! + \! \Re(\tilde{\mathbf{r}})\!^T \mathbf{h} 
\! + \! \Im(\tilde{\mathbf{r}})\!^T \mathbf{g} \Big] \!\! + \!\! ||\mathbf{B}||_F^2, \nonumber
\end{align}
\end{small}
\vspace{-0.3cm}

where,
\begin{small}
\vspace{-0.2cm}
\begin{align}
&\mathbf{M} \!=\!  \text{diag}\big(\!
\sum_{i=0}^{n-1} \Re(\mathcal{F}(\mathbf{x}_i))\!\circ\! \Re(\mathcal{F}(\mathbf{x}_i))
 \!+\!  \Im(\mathcal{F}(\mathbf{x}_i)) \!\circ\! \Im(\mathcal{F}(\mathbf{x}_i))
  \big)  \nonumber\\ 
&\mathbf{h} =  -2 \! \sum_{i=0}^{n-1} 
\Re(\mathcal{F}(\mathbf{x}_i)) \! \circ \!
\Re(\mathcal{F} (\mathbf{B}^T_{i \cdot})) \!
 +  \!
\Im(\mathcal{F}(\mathbf{x}_i)) \circ
\Im(\mathcal{F} (\mathbf{B}^T_{i \cdot})) \nonumber \\ 
&\mathbf{g}  = 2 \sum_{i=0}^{n-1}
\Im(\mathcal{F}(\mathbf{x}_i)) \circ
\Re(\mathcal{F} (\mathbf{B}^T_{i \cdot}))
- 
\Re(\mathcal{F}(\mathbf{x}_i)) \circ
\Im(\mathcal{F} (\mathbf{B}^T_{i \cdot})). \nonumber
\end{align}
\end{small}
%

\vspace{-0.65cm}
For the second term in (\ref{eq:obj}), we note that the circulant matrix can be diagonalized by DFT matrix $\mathbf{F}_d$ and its conjugate transpose $\mathbf{F}_d^H$. Formally, for $\mathbf{R} = \circR(\mathbf{r})$, $\mathbf{r} \in \mathbb{R}^d$,
\begin{equation}
\mathbf{R} = (1/d) \mathbf{F}_d^H \text{diag}(\mathcal{F}(\mathbf{r}))\mathbf{F}_d.
\end{equation}
Let $\Tr(\cdot)$ be the trace of a matrix. Therefore,
\vspace{-0.2cm}
\begin{align}
&||\mathbf{R} \mathbf{R}^T - \mathbf{I}||_F^2 = ||\frac{1}{d} \mathbf{F}_d^H ( \text{diag}(\tilde{\mathbf{r}})^H\text{diag}(\tilde{\mathbf{r}}) - \mathbf{I} )
 \mathbf{F}_d||_F^2 \nonumber\\
  = & \Tr \!\! \left[  \! \frac{1}{d} \mathbf{F}_d^H 
  ( \text{diag}(\tilde{\mathbf{r}})^H\text{diag}(\tilde{\mathbf{r}}) \!\!-\!\! \mathbf{I} )^H
  ( \text{diag}(\tilde{\mathbf{r}})^H\text{diag}(\tilde{\mathbf{r}}) \!\!-\!\! \mathbf{I} )
  \mathbf{F}_d \! \right] \nonumber \\
    = & \Tr \left[   ( \text{diag}(\tilde{\mathbf{r}})^H\text{diag}(\tilde{\mathbf{r}}) - \mathbf{I} )^H
    ( \text{diag}(\tilde{\mathbf{r}})^H\text{diag}(\tilde{\mathbf{r}}) - \mathbf{I} ) \right] \nonumber \\
 = &  || \tilde{\mathbf{r}}^H \circ \tilde{\mathbf{r}} - \mathbf{1} ||_2^2  = || \Re(\tilde{\mathbf{r}})^2 + \Im(\tilde{\mathbf{r}})^2 - \mathbf{1} ||_2^2.
 \label{eq:second}
\end{align}

\vspace{-0.2cm}

Furthermore, as $\mathbf{r}$ is real-valued, additional constraints on $\tilde{\mathbf{r}}$ are needed. For any $u \in \mathbb{C}$, denote $\overline{u}$ as the complex conjugate of $u$. We have the following result \cite{oppenheim1999discrete}: For any real-valued vector $\mathbf{t} \in \mathbb{C}^d$, $\mathcal{F}(\mathbf{t})_0 \text{ is real-valued}$, and
\begin{align}
\mathcal{F}(\mathbf{t})_{d-i} = \overline{ \mathcal{F}(\mathbf{t})_{i} }, \quad\!\! i = 1, \cdots,  \lfloor d/2 \rfloor. \nonumber
\end{align}
\vspace{-0.4cm}

From (\ref{eq:bxr}) $-$ (\ref{eq:second}), the problem of optimizing $\tilde{\mathbf{r}}$ becomes
\begin{align}
\argmin_{\tilde{\mathbf{r}}} \quad\!\! & 
\Re(\tilde{\mathbf{r}})^T \mathbf{M} \Re(\tilde{\mathbf{r}}) + \Im(\tilde{\mathbf{r}})^T \mathbf{M} \Im(\tilde{\mathbf{r}}) 
+ \Re(\tilde{\mathbf{r}})^T \mathbf{h} \nonumber
\\& + \Im(\tilde{\mathbf{r}})^T \mathbf{g} 
 + \lambda d || \Re(\tilde{\mathbf{r}})^2 + \Im(\tilde{\mathbf{r}})^2 - \mathbf{1} ||_2^2 \\
\text{s.t.} \quad &  \Im(\tilde{r}_0) = 0  \nonumber\\
&             \Re(\tilde{r}_i) = \Re(\tilde{r}_{d-i}), i = 1, \cdots, \lfloor d/2 \rfloor \nonumber\\
&             \Im(\tilde{r}_i) = -\Im(\tilde{r}_{d-i}) , i = 1, \cdots, \lfloor d/2 \rfloor.\nonumber
\end{align}
The above is non-convex. Fortunately, the objective function can be decomposed, such that we can solve two variables at a time. Denote the diagonal vector of the diagonal matrix $\mathbf{M}$ as $\mathbf{m}$. The above optimization can then be decomposed to the following sets of optimizations. 
\begin{align} 
&\argmin_{\tilde{r}_0} \quad\!\!\!\!  m_0 \tilde{r}_0^2
+ h_0 \tilde{r}_0 \!+ \lambda d \left( \tilde{r}_0^2 - 1 \right) ^2\!\!, \text{ s.t. }\! \tilde{r}_0 = \overline{\tilde{r}_0}. \label{eq:0}\\
&\argmin_{\tilde{r}_i} \quad\!\!\! (m_i + m_{d-i}) ( \Re(\tilde{r}_i)^2 + \Im(\tilde{r}_i)^2 )  \label{eq:1} \\[-0.2cm]
& \quad\quad\quad + 2 \lambda d \left( \Re(\tilde{r}_i)^2 + \Im(\tilde{r}_i)^2 - 1 \right) ^2 \nonumber \\
& \quad\quad\quad + (h_i + h_{d-i}) \Re(\tilde{r}_i) + (g_i - g_{d-i}) \Im(\tilde{r}_i), \nonumber\\
& \quad\quad\quad\quad i = 1, \cdots, \lfloor d/2 \rfloor. \nonumber
\end{align}
In (\ref{eq:0}), we need to minimize a $4^{th}$ order polynomial with one variable, with the closed form solution readily available. 
In (\ref{eq:1}), we need to minimize a $4^{th}$ order polynomial with two variables.
Though the closed form solution is hard (requiring solution of a cubic bivariate system), we can find local minima by gradient descent, which can be considered as having constant running time for such small-scale problems.
The overall objective is guaranteed to be non-increasing in each step. In practice,  we can get a good solution with just 5-10 iterations.  In summary, the proposed time-frequency alternating optimization procedure has running time $\mathcal{O}(n d \log d)$. 

\subsection{Learning $k < d$ Bits}
\label{subsec:k}
In the case of learning $k < d$ bits, we need to solve the following optimization problem:
\begin{align}
\vspace{-0.5cm} 
\hspace{-0.2cm} 
\argmin_{\mathbf{B}, \mathbf{r}} \quad \!\! & || \mathbf{B}\mathbf{P}_k  \!-\! \mathbf{X} \mathbf{P}_k^T\mathbf{R}^T  ||_F^2 \!+\!  \lambda|| \mathbf{R}\mathbf{P}_k \mathbf{P}_k^T \mathbf{R}^T \!-\! \mathbf{I}||_F^2 \nonumber \\
\text{s.t.}   \quad &  \mathbf{R} = \circR(\mathbf{r}), 
\end{align}
in which 
$\mathbf{P}_k =  
\begin{bmatrix}
  \mathbf{I}_k & \mathbf{O} \\
  \mathbf{O}   & \mathbf{O}_{d-k}
 \end{bmatrix}$,
$\mathbf{I}_k$ is a $k \times k$ identity matrix, and 
$\mathbf{O}_{d-k}$ is a  $(d-k) \times (d-k)$ all-zero matrix.

In fact, the right multiplication of $\mathbf{P}_k$  can be understood as a ``temporal cut-off'', which is equivalent to a
frequency domain convolution. This makes the optimization difficult, as the objective in frequency domain can no longer be decomposed. 
To address this issues, we propose a simple solution in which $B_{ij} = 0$, $i = 0, \cdots, n-1, j = k, \cdots, d-1$ in (\ref{eq:obj}). 
Thus, the optimization procedure remains the same, and the cost is also $\mathcal{O}(n d \log d)$. We will show in experiments that this heuristic provides good performance in practice.

\vspace{-0.2cm}

\section{Experiments}
\label{sec:exp}

\vspace{-0.1cm}

To compare the performance of the proposed circulant binary embedding technique, we conducted experiments on three real-world high-dimensional datasets used by the current state-of-the-art method for generating long binary codes \cite{gonglearning}.  The Flickr-25600 dataset contains 100K images sampled from a noisy Internet image collection. Each image is represented by a $25,600$ dimensional vector.  The ImageNet-51200 contains 100k images sampled from 100 random classes from ImageNet \cite{deng2009imagenet}, each represented by a $51,200$ dimensional vector. The third dataset (ImageNet-25600) is another random subset of ImageNet containing 100K images in $25,600$ dimensional space. All the vectors are normalized to be of unit norm. 

\vspace{-0.1cm}

We compared the performance of the randomized (CBE-rand) and learned (CBE-opt) versions of our circulant embeddings with the current state-of-the-art for high-dimensional data,  \emph{i.e.}, bilinear embeddings. We use both the randomized (bilinear-rand) and learned  (bilinear-opt) versions. Bilinear embeddings have been shown to perform similar or better than another promising technique called Product Quantization \cite{jegou2011product}. Finally, we also compare against the binary codes produced by the baseline LSH method \cite{charikar2002similarity}, which is still applicable to 25,600 and 51,200 dimensional feature but with much longer running time and much more space. We also show an experiment with relatively low-dimensional data in $2048$ dimensional space using Flickr data to compare against techniques that perform well for low-dimensional data but do not scale to high-dimensional scenario. Example techniques include ITQ \cite{gongiterative}, SH \cite{weiss2008spectral}, SKLSH \cite{raginsky2009locality}, and AQBC \cite{gong2012angular}. 

\vspace{-0.1cm}
 
Following \cite{gonglearning, Norouzi11, gordo2011asymmetric}, we use 10,000 randomly sampled instances for training. We then randomly sample 500 instances, different from the training set as queries. The performance (recall@1-100) is evaluated by averaging the recalls of the query instances.
The ground-truth of each query instance is defined as its 10 nearest neighbors based on  $\ell_2$ distance. For each dataset, we conduct two sets of experiments: \textit{fixed-time} where code generation time is fixed and \textit{fixed-bits} where the number of bits is fixed across all techniques. We also show an experiment where the binary codes are used for classification.

\vspace{-0.1cm}

The proposed CBE method is found robust to the choice of $\lambda$ in (\ref{eq:obj}). For example, in the retrieval experiments, the performance difference for $\lambda$ = 0.1, 1, 10, is within 0.5\%. Therefore, in all the experiments, we simply fix $\lambda$ = 1. For the bilinear method, in order to get fast computation, the feature vector is reshaped to a near-square matrix, and the dimension of the two bilinear projection matrices are also chosen to be close to square. Parameters for other techniques are tuned to give the best results on these datasets.

\begin{table}
\begin{small}
\center
\begin{tabular}{|l|l|l|l|}
\hline $d$ & Full proj.   & Bilinear proj. & Circulant proj. \\ 
\hline $2^{15}$ &  $5.44 \times 10^2$ & $2.85$   & $1.11$ \\ 
\hline $2^{17}$ &  -      & $1.91 \times 10^1$  & $4.23$ \\ 
\hline $2^{20}$ (1M) &   -     & $3.76 \times 10^2$ & $3.77 \times 10^1$ \\ 
\hline $2^{24}$ &   -     & $1.22 \times 10^4$     & $8.10 \times 10^2$ \\ 
\hline $2^{27}$ (100M)    &   -    & $2.68\times 10^5$ & $8.15 \times 10^3$ \\ 
\hline 
\end{tabular}
\end{small}
\vspace{-0.2cm}
\caption{Computational time (ms) of full projection (LSH, ITQ, SH \emph{etc}.), bilinear projection (Bilinear), and circulant projection (CBE). The time is based on a single 2.9GHz CPU core. The error is within 10\%. An empty cell indicates that the memory needed for that method is larger than the machine limit of 24GB. }
\label{table:time}
\vspace{-0.4cm}
\end{table}

\textbf{Computational Time.} 
When generating $k$-bit code for $d$-dimensional data, the full projection, bilinear projection, and circulant projection methods have time complexity $O(kd)$, $O(\sqrt{k} d)$, and $O(d \log d)$, respectively. We compare the computational time in Table \ref{table:time} on a fixed hardware. 
Based on our implementation, the computational time of the above three methods can be roughly characterized as $d^2: d\sqrt{d}: 5d\log d $.  Note that faster implementation of FFT algorithms will lead to better computational time for CBE by further reducing the constant factor. Due to the small storage requirement $\mathcal{O}(d)$, and the wide availability of highly optimized FFT libraries, CBE is also suitable for implementation on GPU. Our preliminary tests based on GPU shows up to 20 times speedup compared to CPU. In this paper, for fair comparison, we use same CPU based implementation for all the methods.

\vspace{-0.1cm}

\begin{figure*}[!ht]
\vspace{-0.3cm}
\centering
\subfigure[\# bits (CBE) = 3,200]
{\includegraphics[width = 4.4cm]{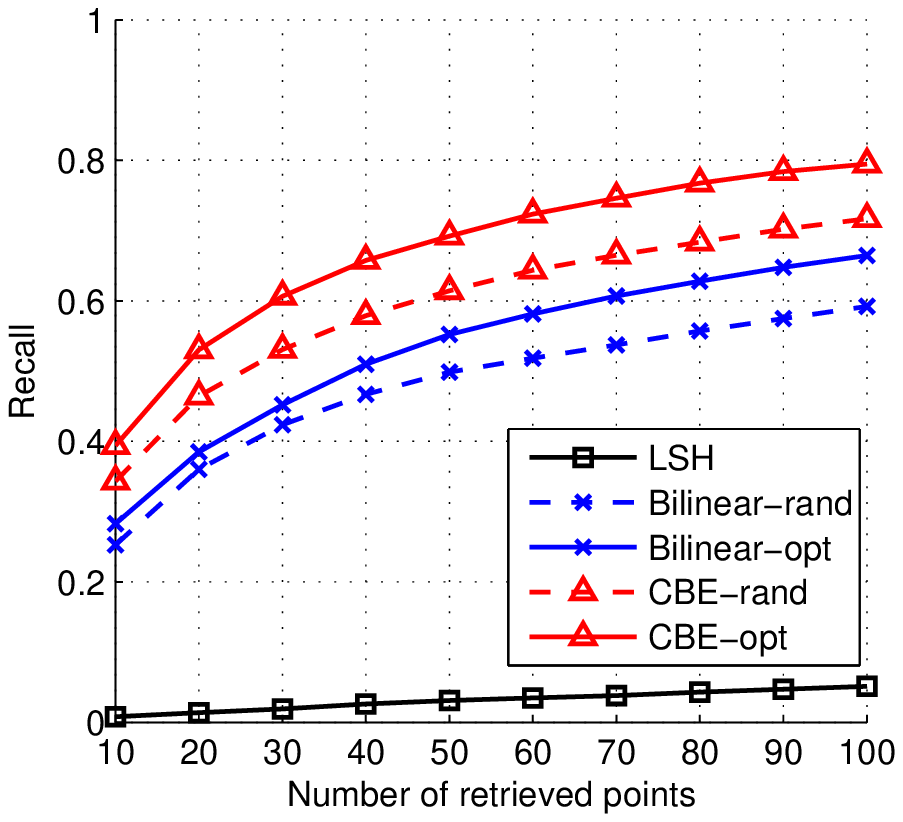}}
\hspace{-0.4cm}
\subfigure[\# bits (CBE) = 6,400]
{\includegraphics[width = 4.4cm]{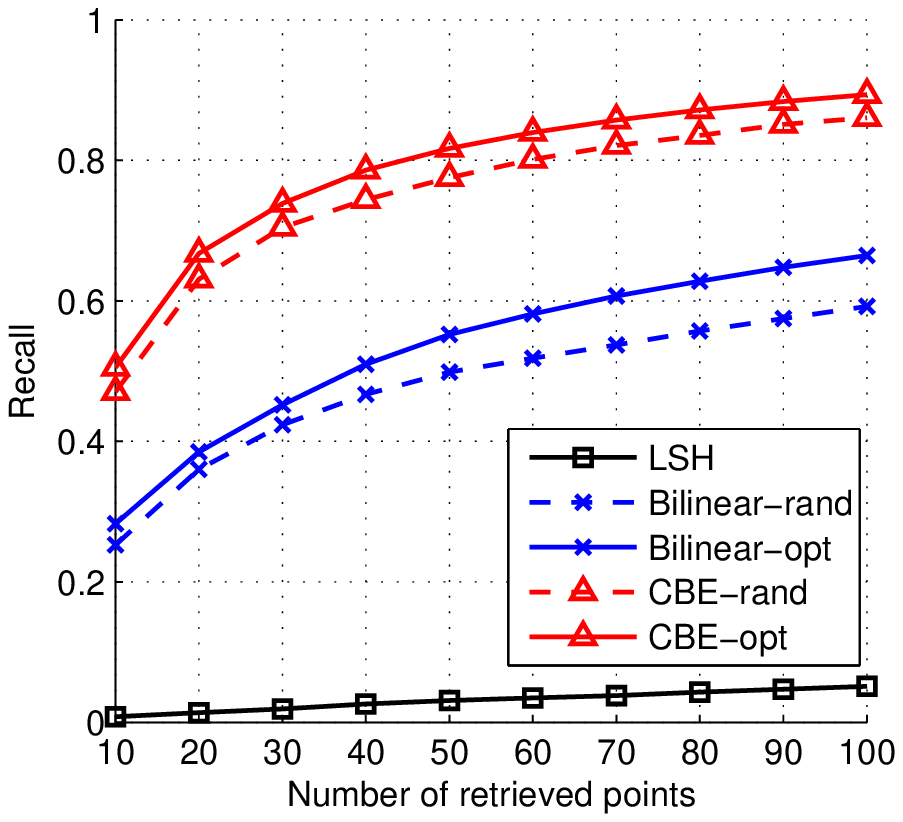}}
\hspace{-0.4cm}
\subfigure[\# bits (CBE) = 12,800]
{\includegraphics[width = 4.4cm]{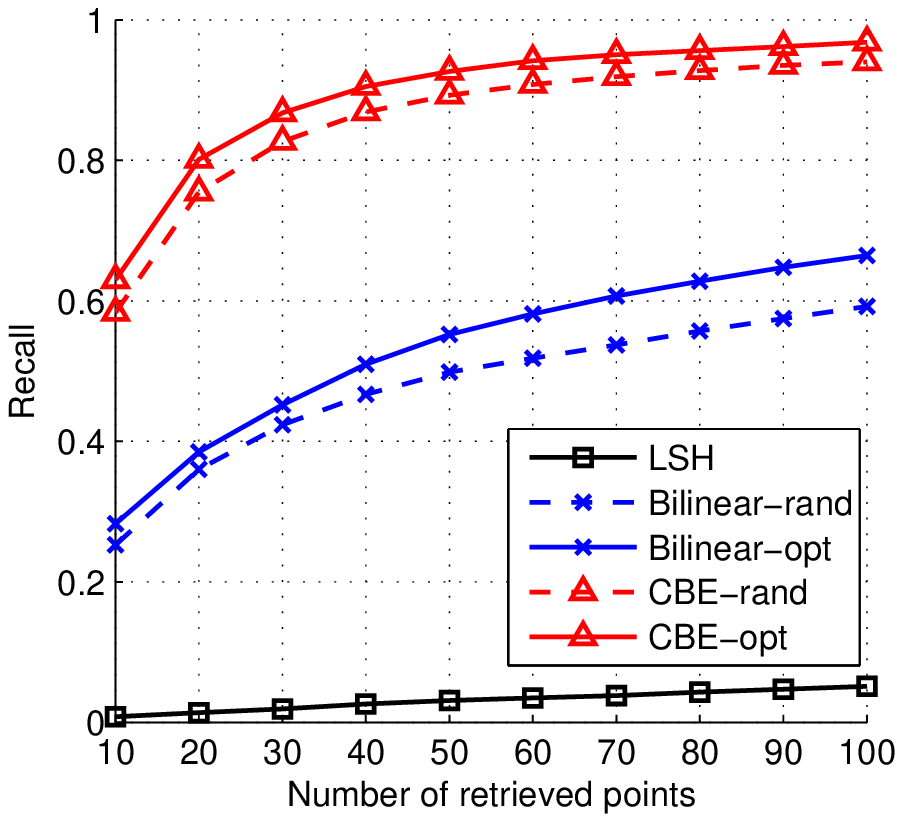}}
\hspace{-0.4cm}
\subfigure[\# bits (CBE) = 25,600]
{\includegraphics[width = 4.4cm]{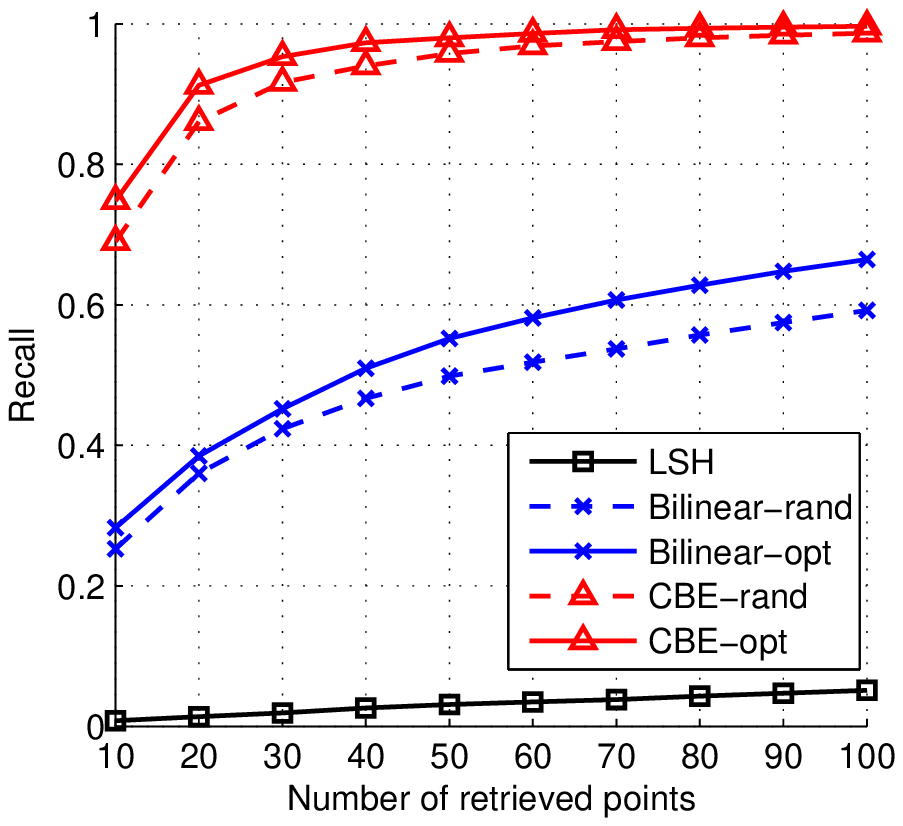}}\\
\vspace{-0.4cm}
\subfigure[\# bits (all) = 3,200]
{\includegraphics[width = 4.4cm]{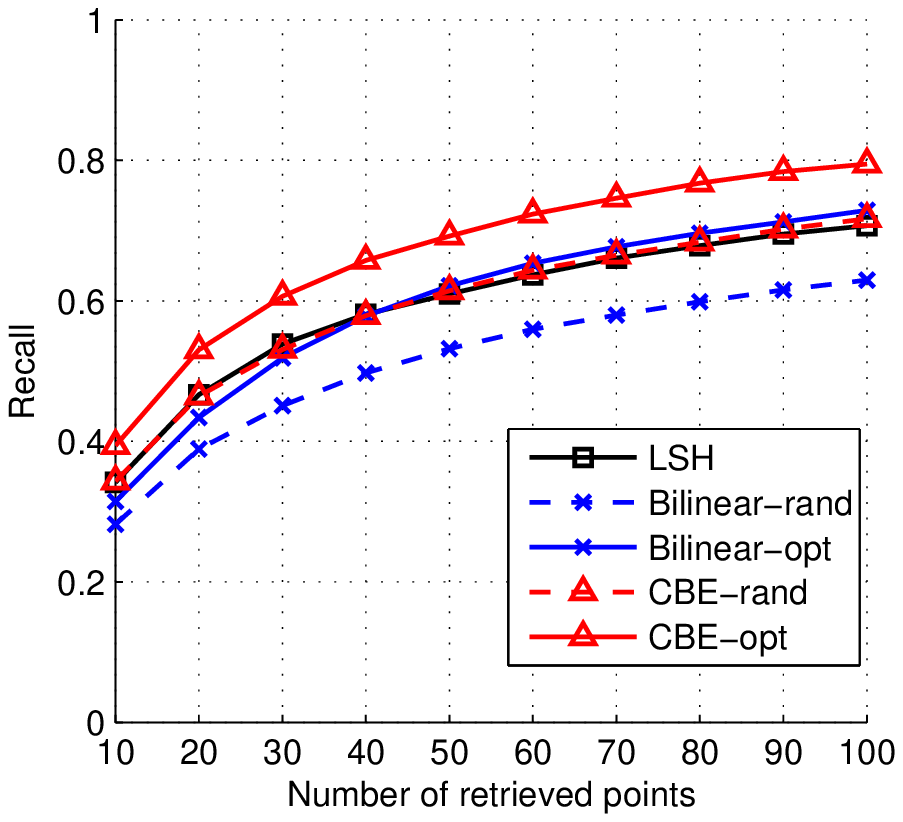}}
\hspace{-0.4cm}
\subfigure[\# bits (all) = 6,400]
{\includegraphics[width = 4.4cm]{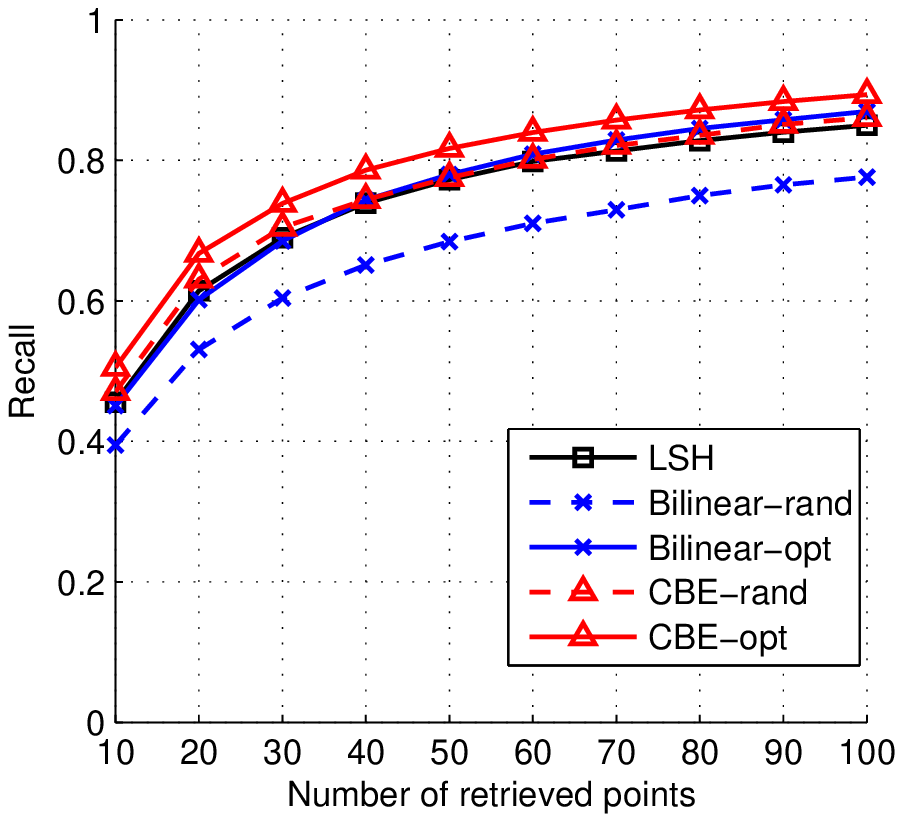}}
\hspace{-0.4cm}
\subfigure[\# bits (all) = 12,800]
{\includegraphics[width = 4.4cm]{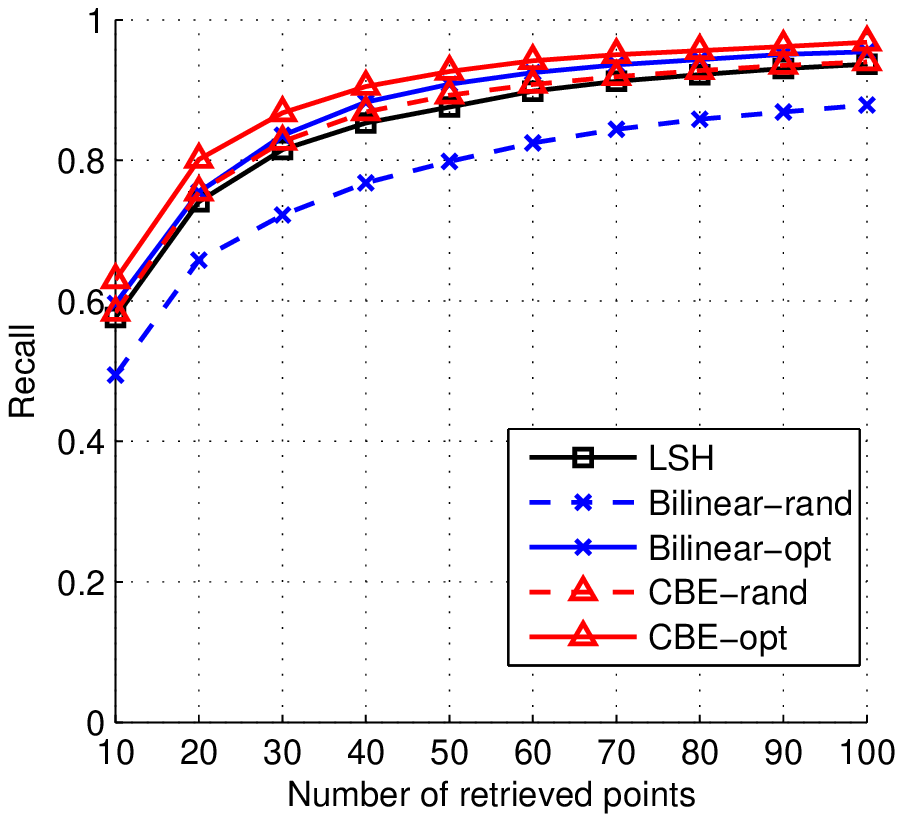}}
\hspace{-0.4cm}
\subfigure[\# bits (all) = 25,600]
{\includegraphics[width = 4.4cm]{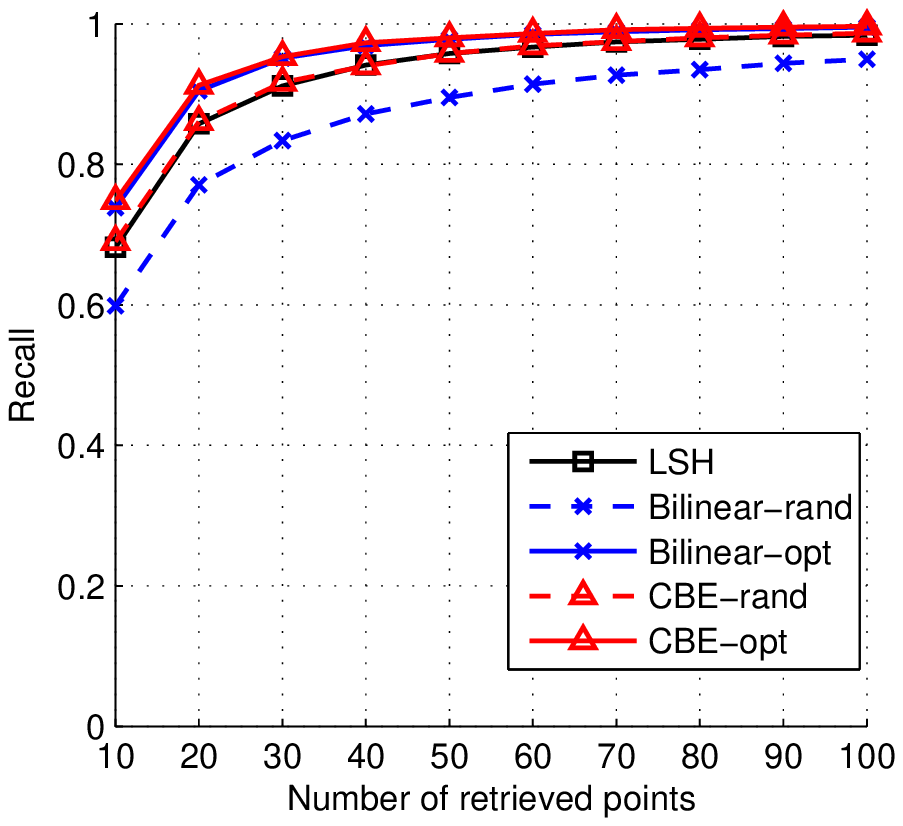}}
\vspace{-0.3cm}
\caption{Recall on Flickr-25600. The standard deviation is within 1\%. \textbf{First Row}: Fixed time. ``\# bits'' is the number of bits of CBE. Other methods are using less bits to make their computational time identical to CBE. 
\textbf{Second Row}: Fixed number of bits. CBE-opt/CBE-rand are 2-3 times faster than Bilinear-opt/Bilinear-rand, and hundreds of times faster than LSH.}
\label{fig:flickr}
\vspace{-0.4cm}
\end{figure*}

\begin{figure*}[!ht]
\centering
\subfigure[\# bits (CBE) = 3,200]
{\includegraphics[width = 4.45cm]{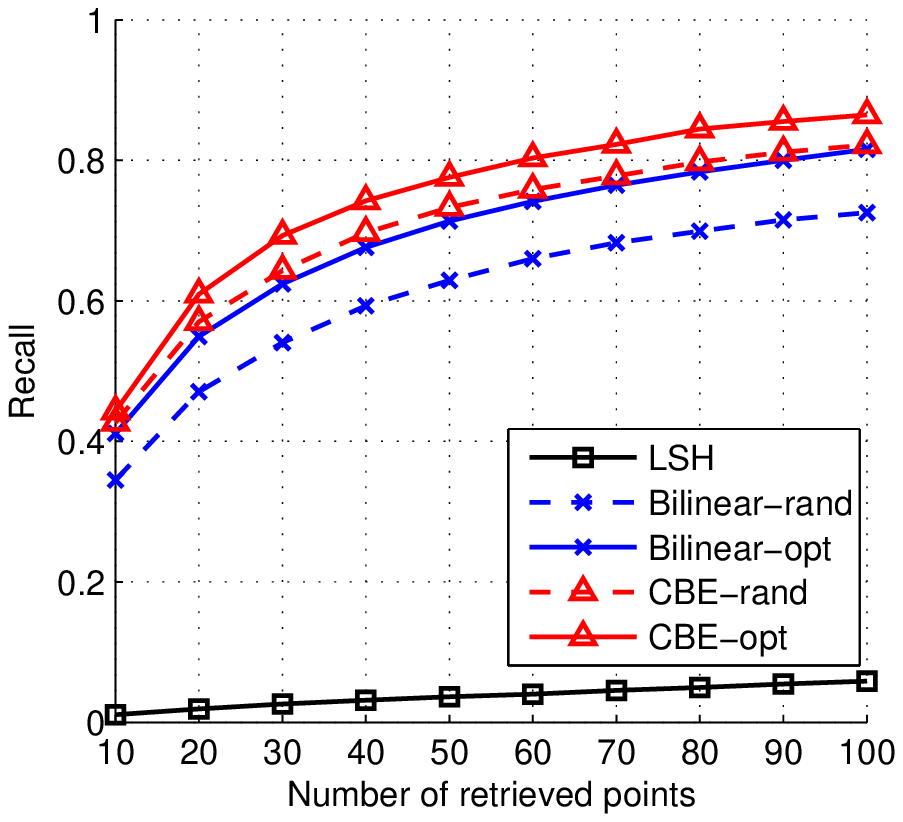}}\hspace{-0.4cm}
\subfigure[\# bits (CBE) = 6,400]
{\includegraphics[width = 4.45cm]{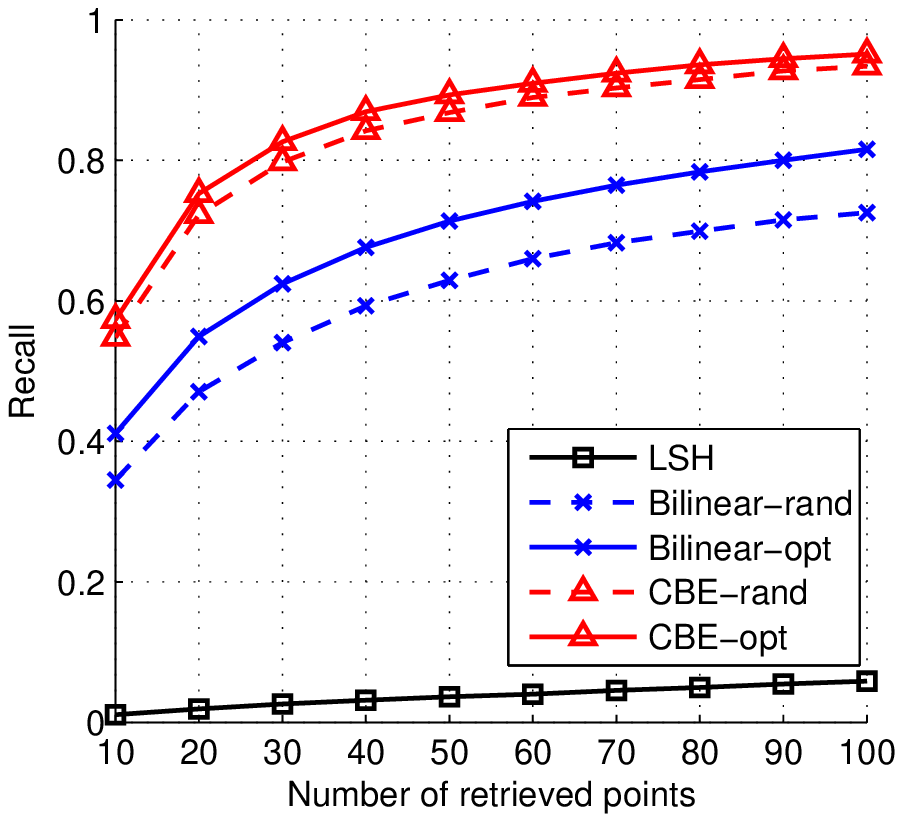}}\hspace{-0.4cm}
\subfigure[\# bits (CBE) = 12,800]
{\includegraphics[width = 4.45cm]{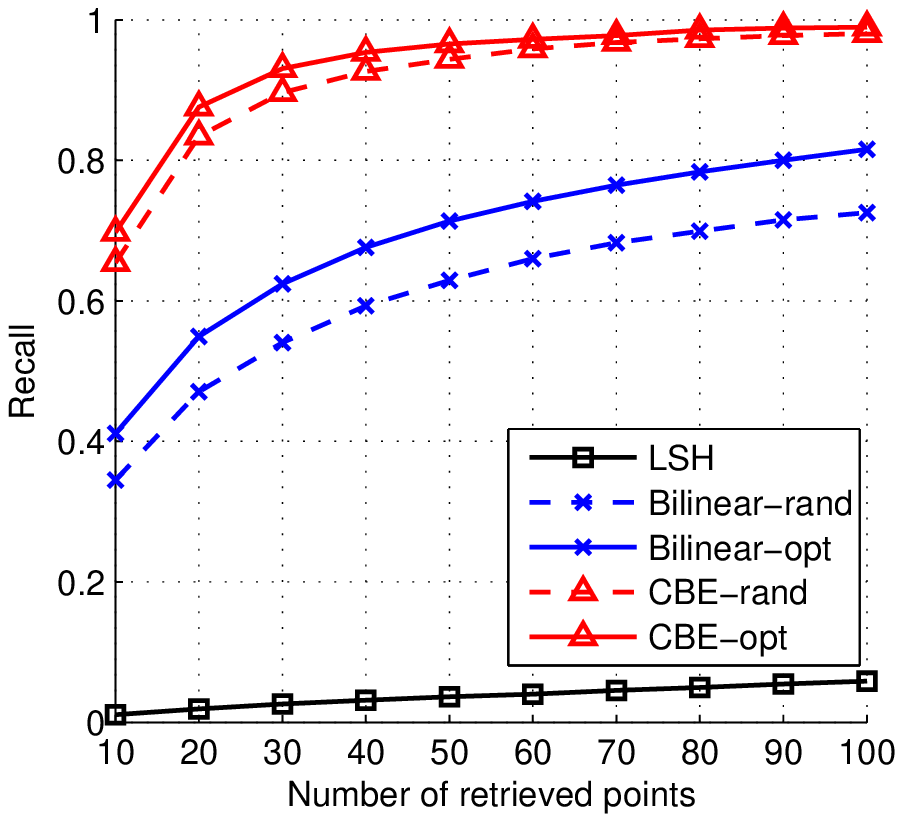}}\hspace{-0.4cm}
\subfigure[\# bits (CBE) = 25,600]
{\includegraphics[width = 4.45cm]{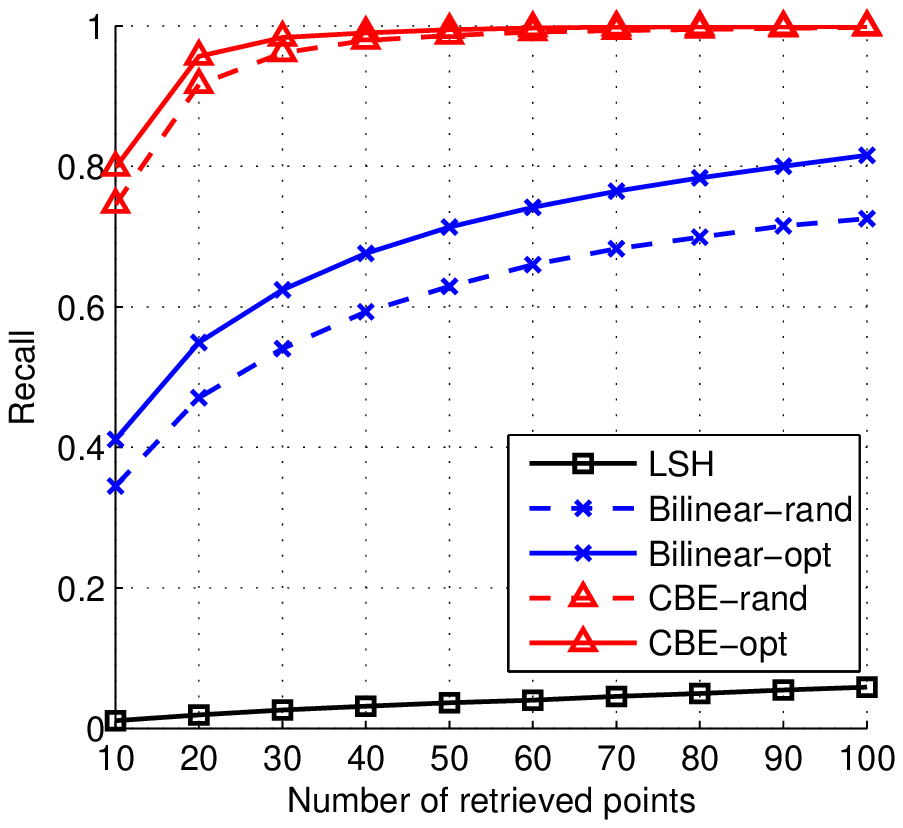}}\\
\vspace{-0.4cm}
\subfigure[\# bits (all) = 3,200]
{\includegraphics[width = 4.45cm]{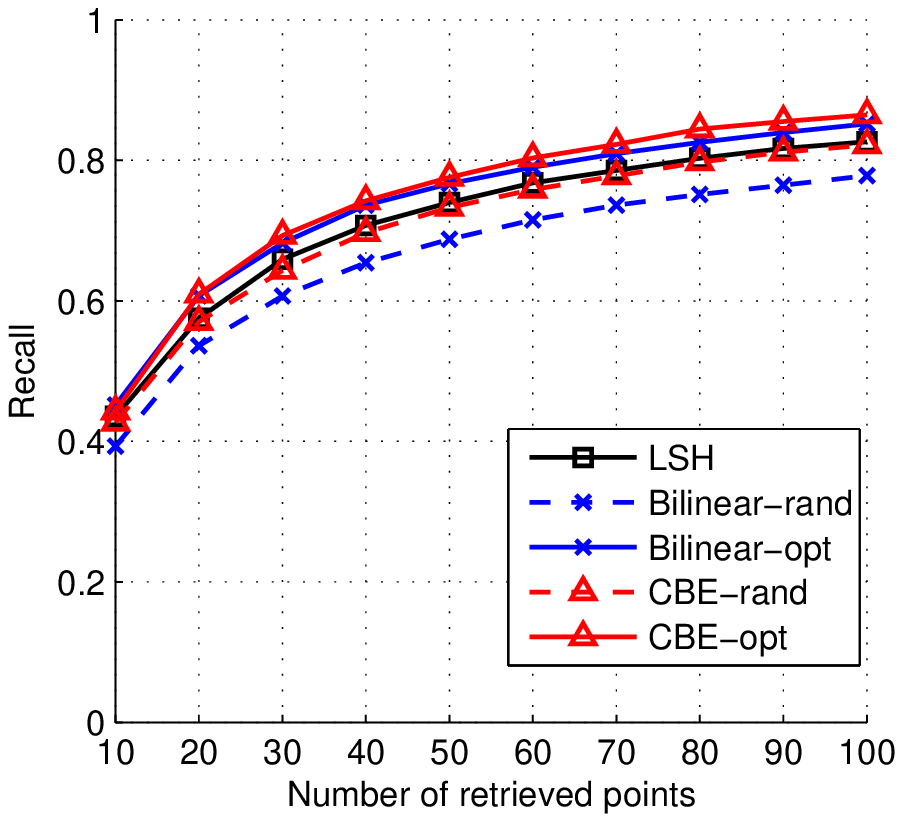}}\hspace{-0.4cm}
\subfigure[\# bits (all) = 64,00]
{\includegraphics[width = 4.45cm]{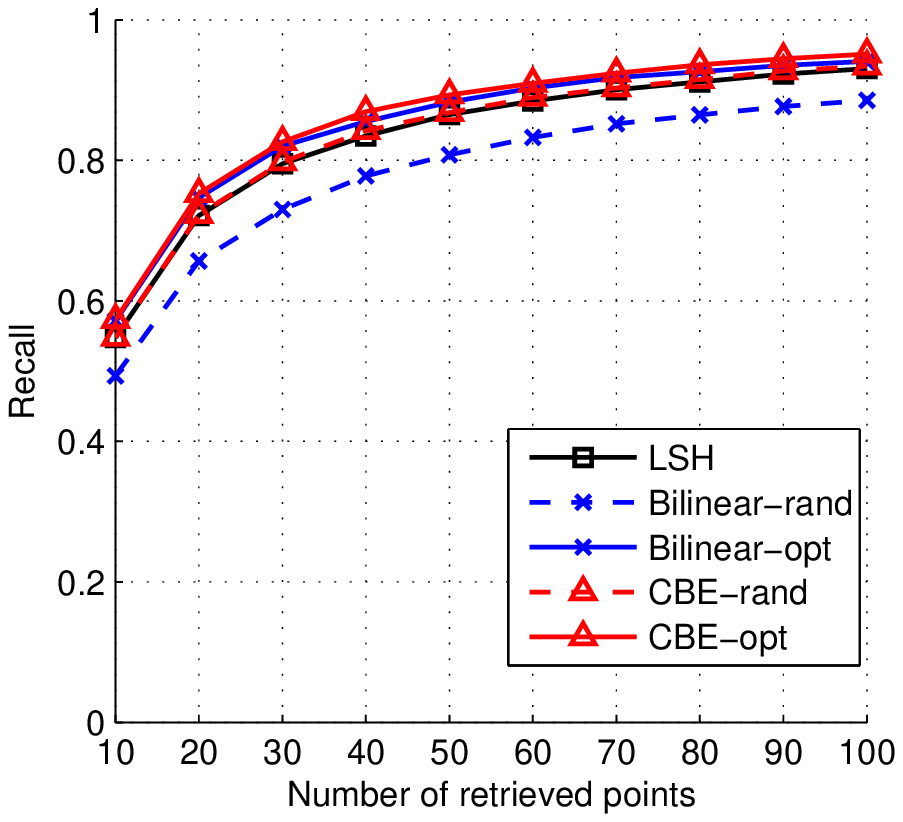}}\hspace{-0.4cm}
\subfigure[\# bits (all) = 12,800]
{\includegraphics[width = 4.45cm]{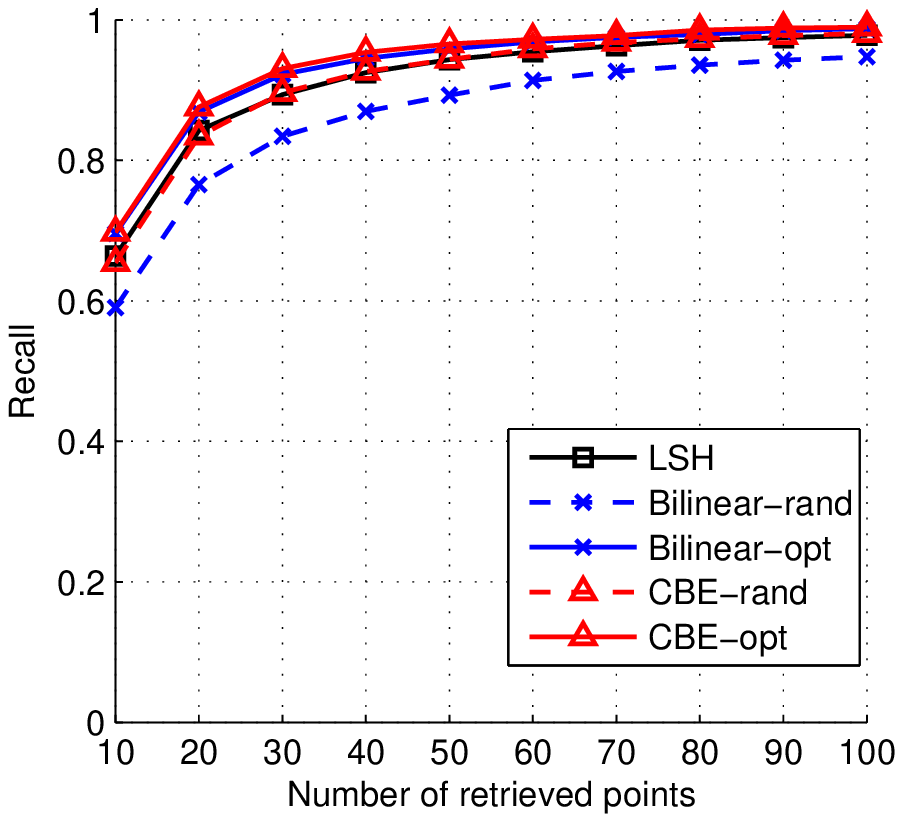}}\hspace{-0.4cm}
\subfigure[\# bits (all) = 25,600]
{\includegraphics[width = 4.45cm]{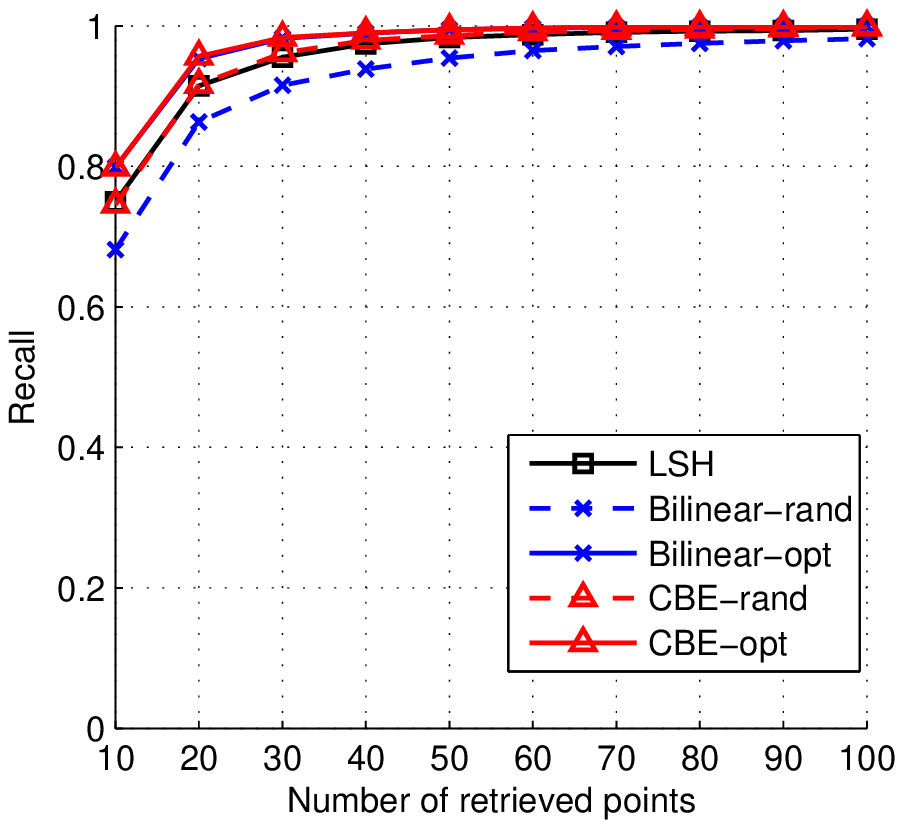}}
\vspace{-0.3cm}
\caption{Recall on ImageNet-25600. The standard deviation is within 1\%. \textbf{First Row}: Fixed time. ``\# bits'' is the number of bits of CBE. Other methods are using less bits to make their computational time identical to CBE. 
\textbf{Second Row}: Fixed number of bits. CBE-opt/CBE-rand are 2-3 times faster than Bilinear-opt/Bilinear-rand, and hundreds of times faster than LSH.}
\label{fig:imagenet}
\vspace{-0.3cm}
\end{figure*}

\begin{figure*}[!ht]
\vspace{-0.3cm}
\centering
\subfigure[\# bits (CBE) = 6,400]
{\includegraphics[width = 4.4cm]{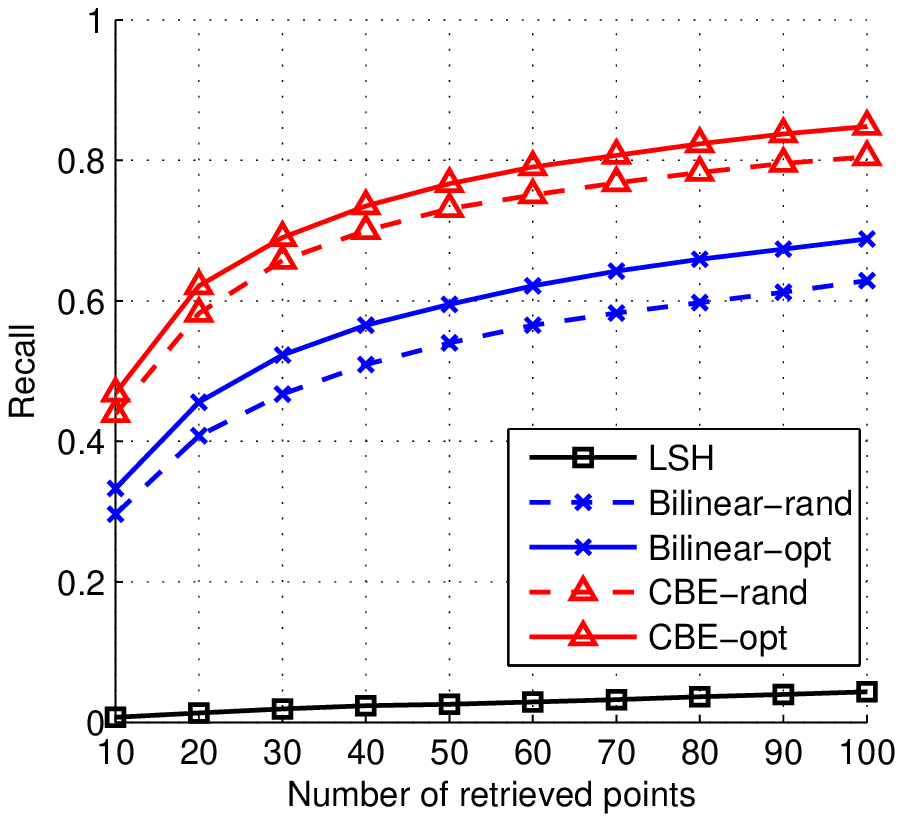}}\hspace{-0.4cm}
\subfigure[\# bits (CBE) = 12,800]
{\includegraphics[width = 4.4cm]{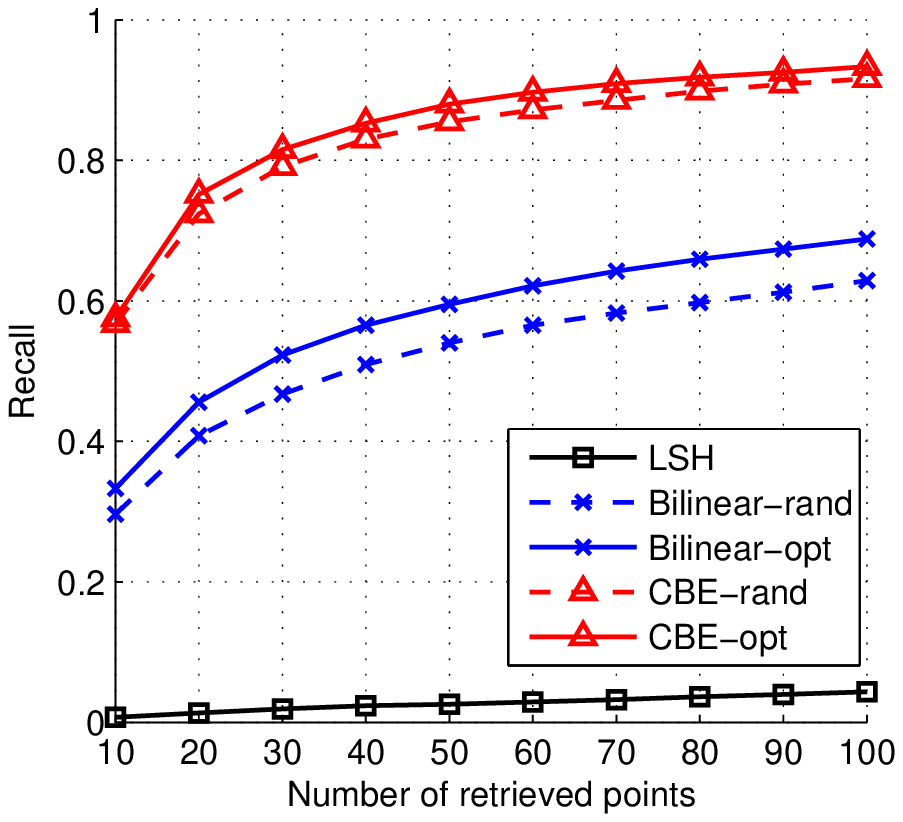}}\hspace{-0.4cm}
\subfigure[\# bits (CBE) = 25,600]
{\includegraphics[width = 4.4cm]{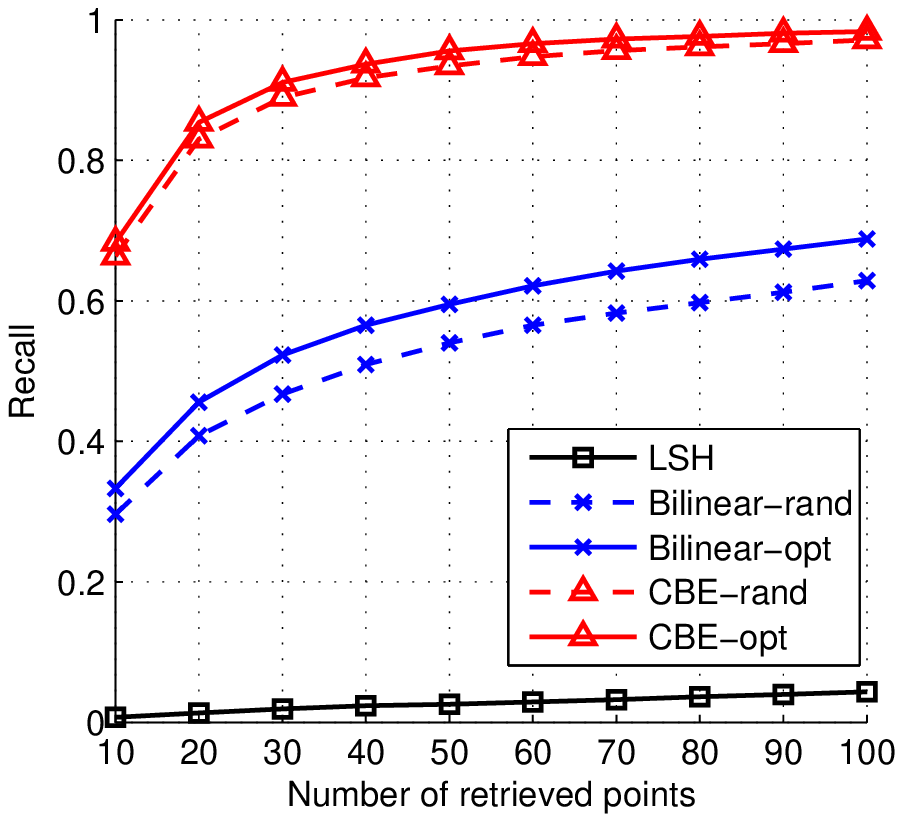}}\hspace{-0.4cm}
\subfigure[\# bits (CBE) = 51,200]
{\includegraphics[width = 4.4cm]{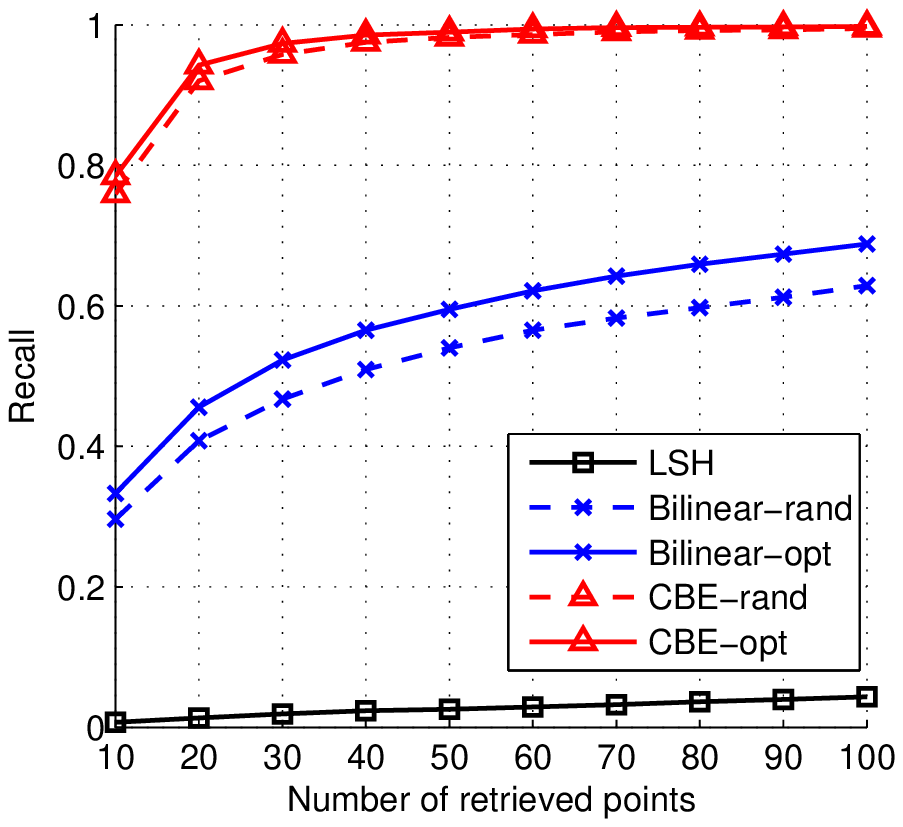}}\\
\vspace{-0.4cm}
\subfigure[\# bits (all) = 6,400]
{\includegraphics[width = 4.4cm]{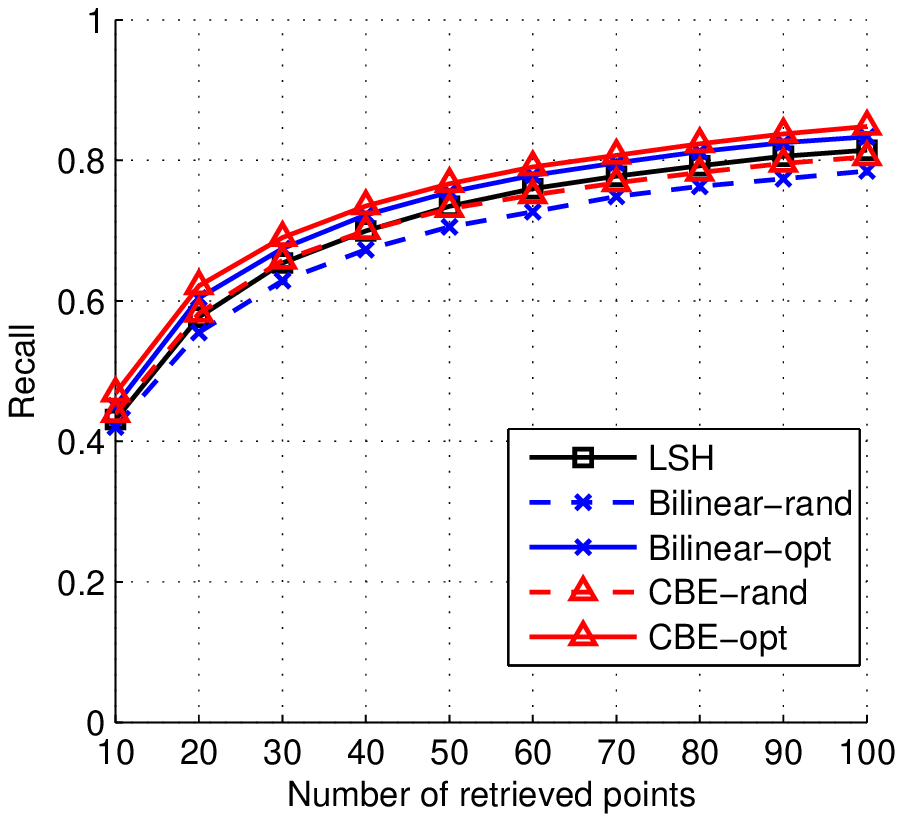}}\hspace{-0.4cm}
\subfigure[\# bits (all) = 12,800]
{\includegraphics[width = 4.4cm]{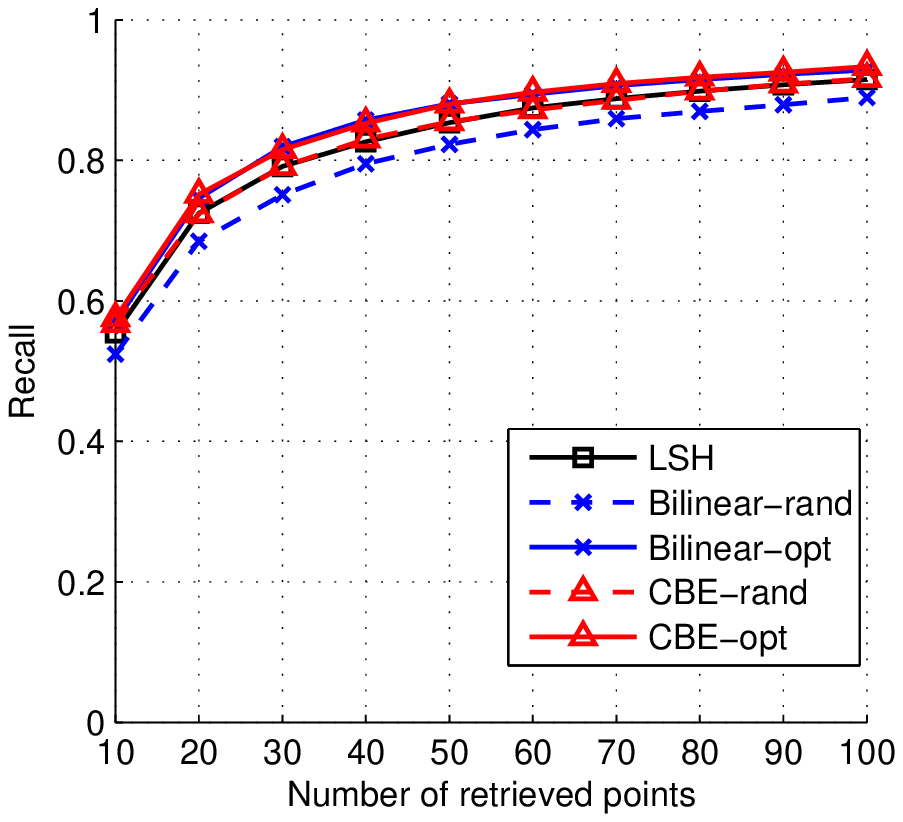}}\hspace{-0.4cm}
\subfigure[\# bits (all) = 25,600]
{\includegraphics[width = 4.4cm]{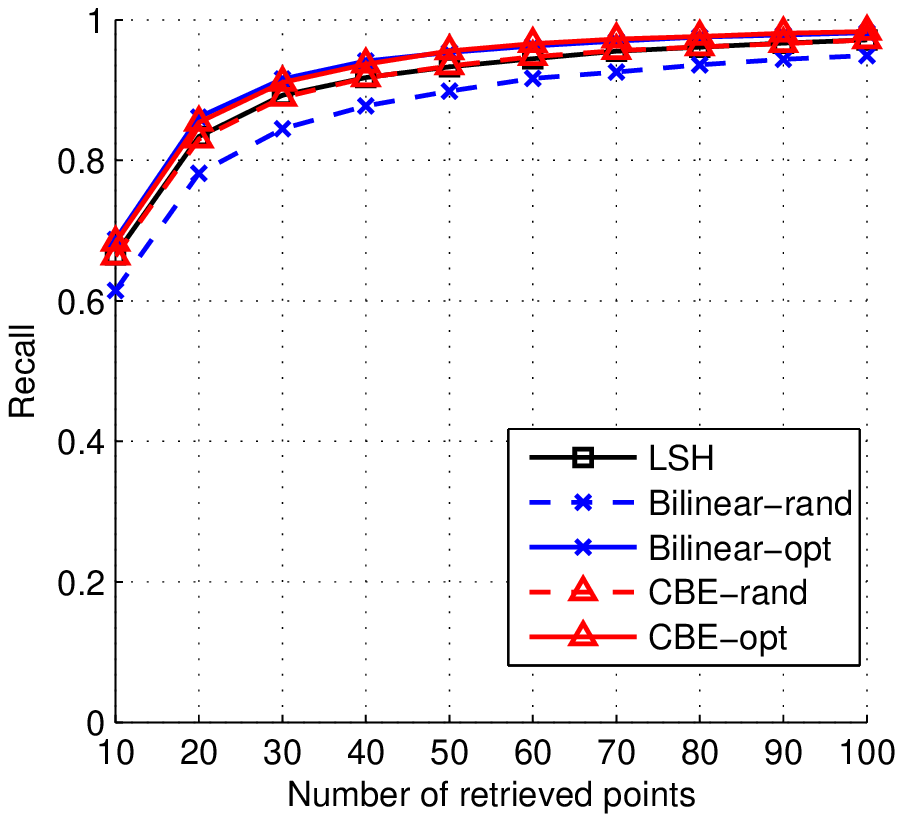}}\hspace{-0.4cm}
\subfigure[\# bits (all) = 51,200]
{\includegraphics[width = 4.4cm]{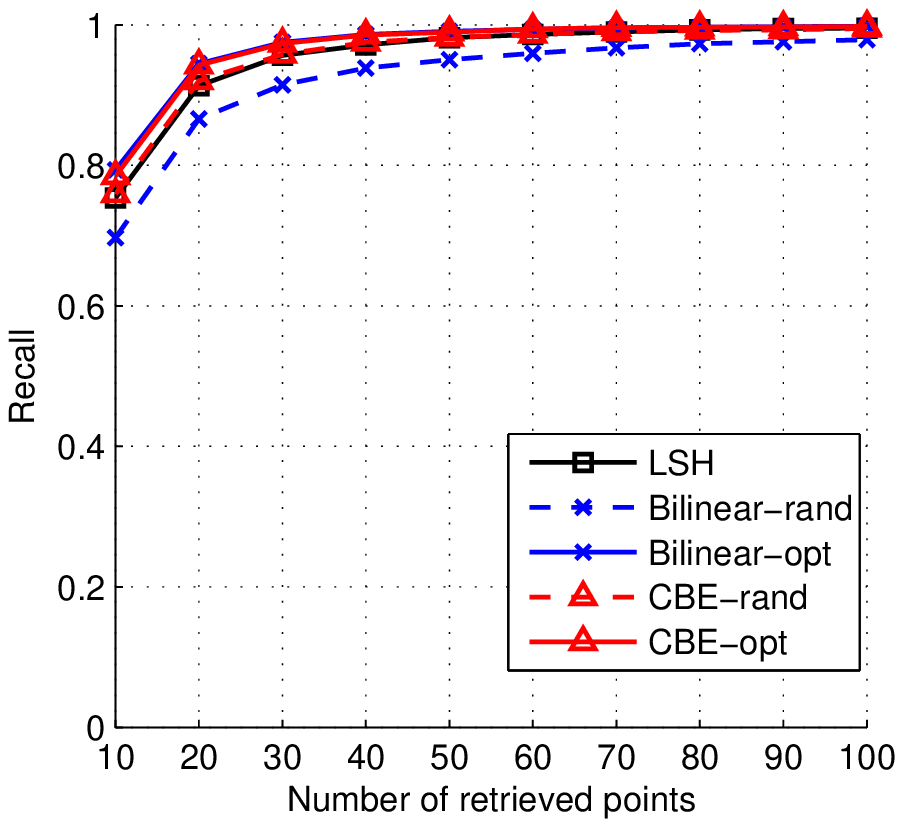}}
\vspace{-0.3cm}
\caption{Recall on ImageNet-51200. The standard deviation is within 1\%. \textbf{First Row}: Fixed time. ``\# bits'' is the number of bits of CBE. Other methods are using less bits to make their computational time identical to CBE. 
\textbf{Second Row}: Fixed number of bits. CBE-opt/CBE-rand are 2-3 times faster than Bilinear-opt/Bilinear-rand, and hundreds of times faster than LSH.}
\label{fig:imagenet_large} 
\end{figure*}

\textbf{Retrieval.} The recall for different methods is compared on the three datasets in 
Figure \ref{fig:flickr}, \ref{fig:imagenet}, and \ref{fig:imagenet_large} respectively. The top row in each figure shows the performance for different methods when the code generation time for all the methods is kept the same as that of CBE. For a fixed time, the proposed CBE yields much better recall than other methods. Even CBE-rand outperforms LSH and Bilinear code by a large margin. The second row compares the performance for different techniques with codes of same length. In this case, the performance of CBE-rand is almost identical to LSH even though it is hundreds of time faster. This is consistent with our analysis in Section \ref{sec:rand}. Moreover, CBE-opt/CBE-rand outperform the Bilinear-opt/Bilinear-rand in addition to being 2-3 times faster.

\vspace{-0.1cm}

\textbf{Classification.} Besides retrieval, we also test the binary codes for classification. The advantage is to save on storage allowing even large scale datasets to fit in memory  \cite{li2011hashing, sanchez2011high}.
We follow the asymmetric setting of \cite{sanchez2011high} by training linear SVM on binary code $\text{sign}(\mathbf{R} \mathbf{x})$, and testing on the original $\mathbf{R} \mathbf{x}$. 
This empirically has been shown to give better accuracy than the symmetric procedure.
We use ImageNet-25600, with randomly sampled 100 images per category for training, 50 for validation and 50 for testing. The code dimension is set as 25,600.
As shown in Table \ref{table:classification}, CBE, which has much faster computation, does not show any performance degradation compared to LSH or bilinear codes in classification task as well.
\vspace{-0.1cm}

\begin{table}[t]
\center
\begin{small}
\begin{tabular}{|c|c|c|c|}
\hline
  Original& LSH  & Bilinear-opt  & CBE-opt  \\ 
\hline  25.59$\pm$0.33 & 23.49$\pm$0.24  & 24.02$\pm$0.35 & 24.55 $\pm$0.30  \\ 
\hline
\end{tabular} 
\end{small}
\caption{Multiclass classification accuracy on binary coded ImageNet-25600. The binary codes of same dimensionality are 32 times more space efficient than the original features (single-float).}
\label{table:classification}
\vspace{-0.4cm}
\end{table}
\begin{figure}[!t]
\centering
\subfigure[\# bits = 1,024]
{\includegraphics[width = 4.2cm]{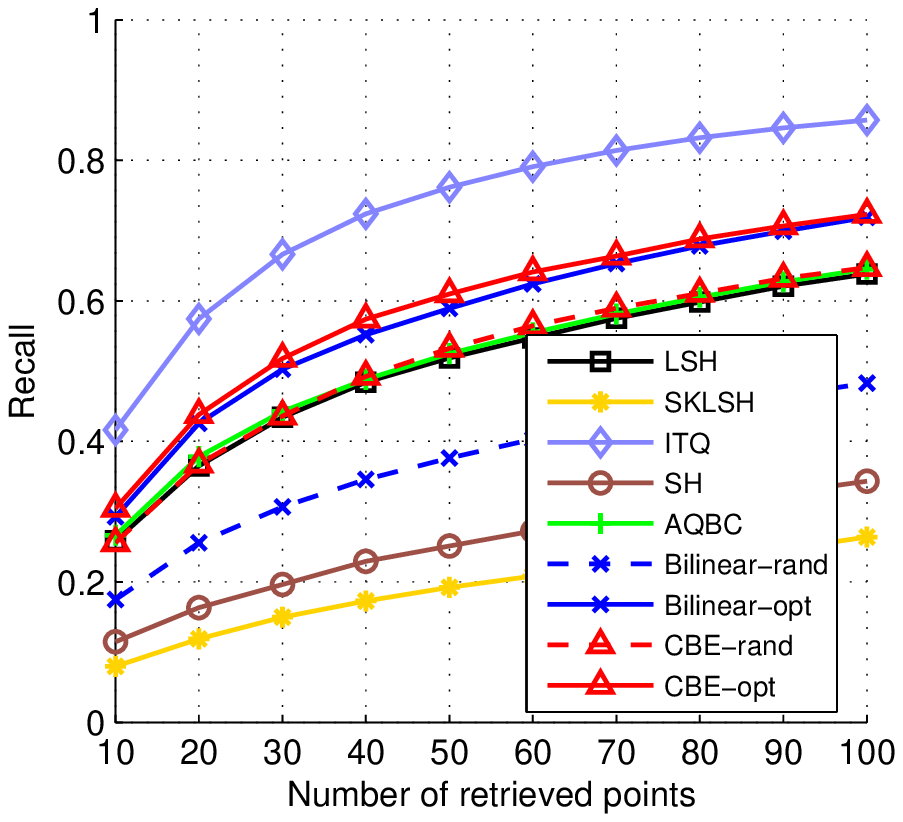}}
\hspace{-0.4cm}
\subfigure[\# bits = 2,048]
{\includegraphics[width = 4.2cm]{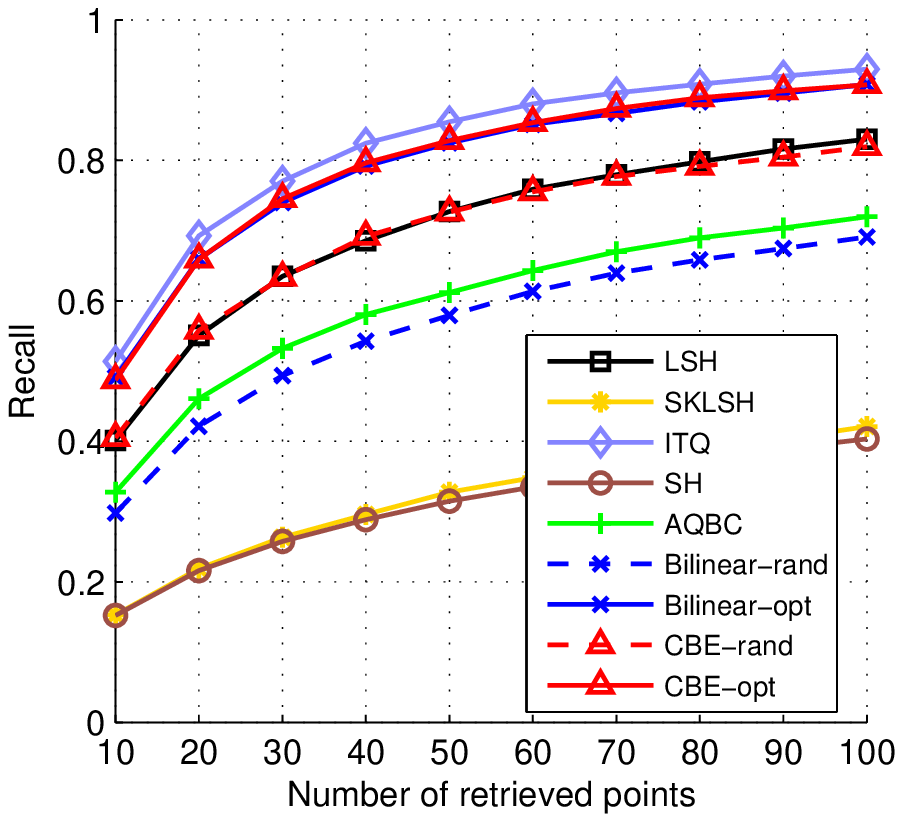}}
\vspace{-0.1cm}
\caption{Performance comparison on relatively low-dimensional data (Flickr-2048) with fixed number of bits. CBE gives comparable performance to the state-of-the-art even on low-dimensional data as the number of bits is increased. However, note that these other methods do not scale to very high-dimensional data setting which is the main focus of this work.}
\label{fig:flickr_2048}
\vspace{-0.4cm}
\end{figure}

\textbf{Low-Dimensional Experiment.}   There exist several techniques that do not scale to high-dimensional case. To compare our method with those, we conducted experiments with fixed number of bits on a relatively low-dimensional dataset (Flickr-2048), constructed by randomly sampling 2,048 dimensions of Flickr-25600. As shown in Figure \ref{fig:flickr_2048}, though CBE is not designed for such scenario, the CBE-opt performs better or equivalent to other techniques except ITQ which scales very poorly with $d$ ($\mathcal{O}(d^3)$). Moreover, as the number of bits increases, the gap between ITQ and CBE becomes much smaller suggesting that the performance of ITQ is not expected to be better than CBE even if one could run ITQ on high-dimensional data.

\section{Semi-supervised Extension}

In some applications, one can have access to a few labeled pairs of similar and dissimilar data points. Here we show how the CBE formulation can be extended to incorporate such information in learning.  This is achieved by adding an additional objective term $J(\mathbf{R})$.
\begin{align}
\argmin_{\mathbf{B}, \mathbf{r}} & || \mathbf{B} \!-\! \mathbf{X} \mathbf{R}^T ||_F^2 
\!+\! \lambda ||\mathbf{R} \mathbf{R}^T - \mathbf{I}||_F^2
+ \mu J(\mathbf{R}) \\
\text{s.t.}  \quad & \mathbf{R} = \circR(\mathbf{r}), \nonumber
\end{align}
\vspace{-0.4cm}
\begin{small}
\begin{align}
J(\mathbf{R}) \!\! =  \!\!& 
\sum_{i,j \in \mathcal{M}}  \!\!||\mathbf{R}\mathbf{x}_i  \!\!- \!\! \mathbf{R}\mathbf{x}_j  ||_2^2
 \!\!-  \!\!
\sum_{i,j \in \mathcal{D}} ||\mathbf{R}\mathbf{x}_i \!\! - \!\! \mathbf{R}\mathbf{x}_j  ||_2^2.
\end{align}
\vspace{-0.4cm}
\end{small}

Here $\mathcal{M}$ and $\mathcal{D}$ are the set of ``similar'' and ``dissimilar'' instances, respectively. The intuition is to maximize the distances between the dissimilar pairs, and minimize the distances between the similar pairs. Such a term is commonly used in semi-supervised binary coding methods \cite{wang2010sequential}. 
We again use the time-frequency alternating optimization procedure of Section \ref{sec:opt}. For a fixed $\mathbf{r}$, the optimization procedure to update $\mathbf{B}$ is the same.  For a fixed $\mathbf{B}$, optimizing $\mathbf{r}$ is done in frequency domain by expanding $J(\mathbf{R})$ as below, with similar techniques used in Section \ref{sec:opt}. 
\begin{align}
\vspace{-0.5cm}
||\mathbf{R}\mathbf{x}_i \!-\! \mathbf{R}\mathbf{x}_j  ||_2^2 = (1/d) || \text{diag}(\mathcal{F}(\mathbf{x}_i)\!-\!\mathcal{F}(\mathbf{x}_j))  \tilde{\mathbf{r}} ||_2^2. \nonumber
\end{align}
\vspace{-0.2cm}
Therefore,
\begin{align}
\vspace{-0.3cm}
J(\mathbf{R}) = (1/d) ( \Re(\tilde{\mathbf{r}})^T \mathbf{A} \Re(\tilde{\mathbf{r}}) + \Im(\tilde{\mathbf{r}})^T \mathbf{A} \Im(\tilde{\mathbf{r}}) ),
\vspace{-0.3cm}
\end{align}
where, $\mathbf{A} =  \mathbf{A}_1 + \mathbf{A}_2 - \mathbf{A}_3  - \mathbf{A}_4 $, and
\begin{small}
\begin{equation}
\mathbf{A}_1 \!\!=\!\!\!\! \sum_{(i,j) \in \mathcal{M}}\!\!\!\! \Re(\text{diag}(\mathcal{F}(\mathbf{x}_i) - \mathcal{F}(\mathbf{x}_j)))^T \Re(\text{diag} (\mathcal{F}(\mathbf{x}_i) - \mathcal{F}(\mathbf{x}_j))),
\nonumber
\end{equation}
\end{small}
\vspace{-0.2cm}
\begin{small}
\begin{equation}
 \mathbf{A}_2 \!\!=\!\!\!\! \sum_{(i,j) \in \mathcal{M}}\!\!
\Im(\text{diag} (\mathcal{F}(\mathbf{x}_i) - \mathcal{F}(\mathbf{x}_j)))^T \Im(\text{diag} (\mathcal{F}(\mathbf{x}_i)-\mathcal{F}(\mathbf{x}_j))),
\nonumber
\end{equation}
\end{small}
\vspace{-0.2cm}
\begin{small}
\begin{equation}
\mathbf{A}_3 \!\!=\!\!\!\! \sum_{(i,j) \in \mathcal{D}}\!\!
\Re(\text{diag}(\mathcal{F}(\mathbf{x}_i) - \mathcal{F}(\mathbf{x}_j)))^T \Re(\text{diag} (\mathcal{F}(\mathbf{x}_i) - \mathcal{F}(\mathbf{x}_j))),
\nonumber
\end{equation}
\end{small}
\vspace{-0.2cm}
\begin{small}
\begin{equation}
\mathbf{A}_4 \!\!=\!\!\!\!  \sum_{(i,j) \in \mathcal{D}}\!\!
\Im(\text{diag} (\mathcal{F}(\mathbf{x}_i) - \mathcal{F}(\mathbf{x}_j)))^T \Im(\text{diag} (\mathcal{F}(\mathbf{x}_i)-\mathcal{F}(\mathbf{x}_j))).
\nonumber
\end{equation}
\end{small}

\vspace{-0.4cm}

Hence, the optimization can be carried out as in Section \ref{sec:opt}, where $\mathbf{M}$ in (\ref{eq:bxr}) is simply replaced by $\mathbf{M} + \mu \mathbf{A}$. Our experiments show that the semi-supervised extension improves over the non-semi-supervised version by 2\% in terms of averaged AUC on the ImageNet-25600 dataset.
\vspace{-0.3cm}

\section{Conclusion}
We have proposed circulant binary embedding for generating long codes for very high-dimensional data. A novel time-frequency alternating optimization was also introduced to learn the model parameters from the training data. The proposed method has time complexity $\mathcal{O}(d\log d)$ and space complexity $\mathcal{O}(d)$, while showing no performance degradation on real-world data compared to more expensive approaches ($\mathcal{O}(d^2)$ or $\mathcal{O}(d^{1.5})$). On the contrary, for the fixed time, it showed significant accuracy gains. The full potential of the method can be unleashed when applied to ultra-high dimensional data (say $d\sim$100M), for which no other methods are applicable. 

\bibliographystyle{icml2014}
\bibliography{design}

\end{document}